\documentclass[10pt,twocolumn,letterpaper]{article}

\usepackage[pagenumbers]{cvpr}

\usepackage[utf8]{inputenc}
\usepackage[T1]{fontenc}

\usepackage{graphicx}
\usepackage[table]{xcolor}
\usepackage{amssymb}
\usepackage{stmaryrd}
\usepackage{pgffor}
\usepackage{pgfmath}

\usepackage{multirow}
\usepackage{booktabs}
\usepackage{adjustbox}
\usepackage[most]{tcolorbox}
\usepackage{wrapfig}
\usepackage{xspace}
\usepackage{pifont}
\usepackage{caption} 

\usepackage{scalerel,stackengine}
\stackMath

\newcommand{\cmark}{\textcolor{green!60!black}{\ding{51}}} 
\newcommand{\xmark}{\textcolor{red}{\ding{55}}}       

\DeclareRobustCommand{\Stars}[1]{%
  \mbox{%
    \ensuremath{%
      \ifcase#1\relax
        \scalerel*{\stackinset{c}{-.125pt}{c}{}{\scalebox{.9}{\textcolor{white}{$\bigstar$}}}{\bigstar}}{\bigstar}%
        \scalerel*{\stackinset{c}{-.125pt}{c}{}{\scalebox{.9}{\textcolor{white}{$\bigstar$}}}{\bigstar}}{\bigstar}%
        \scalerel*{\stackinset{c}{-.125pt}{c}{}{\scalebox{.9}{\textcolor{white}{$\bigstar$}}}{\bigstar}}{\bigstar}%
      \or
        \bigstar%
        \scalerel*{\stackinset{c}{-.125pt}{c}{}{\scalebox{.9}{\textcolor{white}{$\bigstar$}}}{\bigstar}}{\bigstar}%
        \scalerel*{\stackinset{c}{-.125pt}{c}{}{\scalebox{.9}{\textcolor{white}{$\bigstar$}}}{\bigstar}}{\bigstar}%
      \or
        \bigstar\bigstar%
        \scalerel*{\stackinset{c}{-.125pt}{c}{}{\scalebox{.9}{\textcolor{white}{$\bigstar$}}}{\bigstar}}{\bigstar}%
      \or
        \bigstar\bigstar\bigstar%
      \fi
    }%
  }%
}

\makeatletter
\DeclareRobustCommand\onedot{\futurelet\@let@token\@onedot}
\def\@onedot{\ifx\@let@token.\else.\null\fi\xspace}
\makeatother

\newcommand{\app}{\raise.17ex\hbox{$\scriptstyle\sim$}}

\definecolor{cvprblue}{rgb}{0.21,0.49,0.74}
\usepackage[pagebackref,breaklinks,colorlinks,allcolors=cvprblue]{hyperref}

\author{\textbf{Chenhui Gou$^{1,4*\dagger}$} \quad 
\textbf{Zilong Chen$^{2,4*\dagger}$} \quad 
\textbf{Zeyu Wang$^{3*}$} \quad 
\textbf{Feng Li$^{4}$} \quad 
\textbf{Deyao Zhu$^{4}$}\\
\textbf{Zicheng Duan}$^{5}$ \quad 
\textbf{Kunchang Li}$^{4}$ \quad 
\textbf{Chaorui Deng}$^{4}$ \quad 
\textbf{Hongyi Yuan}$^{4}$\\
\textbf{\textbf{Haoqi Fan}$^{4}$} \quad 
\textbf{Cihang Xie}$^{3}$ \quad 
\textbf{Jianfei Cai}$^{1}$ \quad 
\textbf{Hamid Rezatofighi}$^{1}$\vspace{.3em}\\
$^{1}$Monash University \quad 
$^{2}$Tsinghua University \quad 
$^{3}$UC Santa Cruz \\
$^{4}$Bytedance Seed \quad 
$^{5}$University of Adelaide\vspace{.3em}
  \\  
  \small \centerline{\includegraphics[height=1.1em]{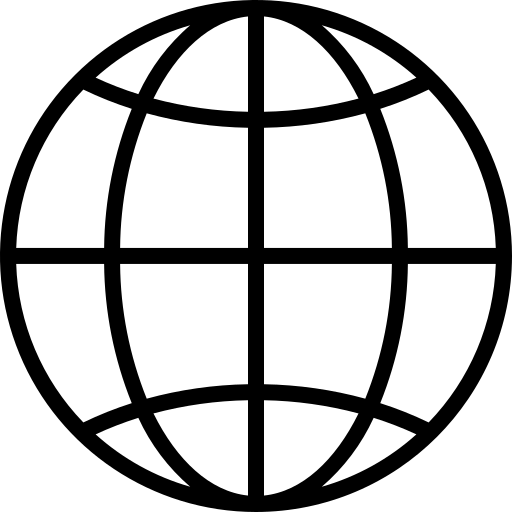} \textbf{Project Page}: \url{https://chenhuigou.github.io/VQ-VA-World/}}
  }
\title{VQ-VA World: Towards High-Quality Visual Question-Visual Answering}

\begin{document}
\maketitle

\begingroup
\renewcommand\thefootnote{*}
\footnotetext{Equal contribution.}
\renewcommand\thefootnote{$\dagger$}
\footnotetext{Work done during internship}
\endgroup

\begin{abstract}
This paper studies \textit{Visual Question-Visual Answering (VQ-VA)}: generating an image, rather than text, in response to a visual question---an ability that has recently emerged in proprietary systems such as NanoBanana and GPT-Image. 
To also bring this capability to open-source models, we introduce VQ-VA World, a data-centric framework built around an agentic pipeline for large-scale, targeted data construction. 
Leveraging web-scale deployment, this pipeline crawls a massive amount of ~1.8M high-quality, interleaved image-text samples for model training. For evaluation, we further release IntelligentBench, a human-curated benchmark that systematically assesses VQ-VA along the aspects of \textit{world knowledge}, \textit{design knowledge}, and \textit{reasoning}. 
Training with VQ-VA World data yields strong empirical gains: it helps LightFusion attain 53.06 on IntelligentBench, substantially surpassing the best prior open-source baselines (\emph{i.e.}, 7.78 from vanilla LightFusion; 1.94 from UniWorld-V1), and significantly narrowing the gap toward leading proprietary systems (\emph{e.g.}, 81.67 from NanoBanana; 82.64 from GPT-Image). By releasing the full suite of model weights, datasets, and pipelines, we hope to stimulate future research on VQ-VA.

\end{abstract}
\section{Introduction}
\begin{figure*}[t]
    \centering
    \includegraphics[width=1.0\linewidth]{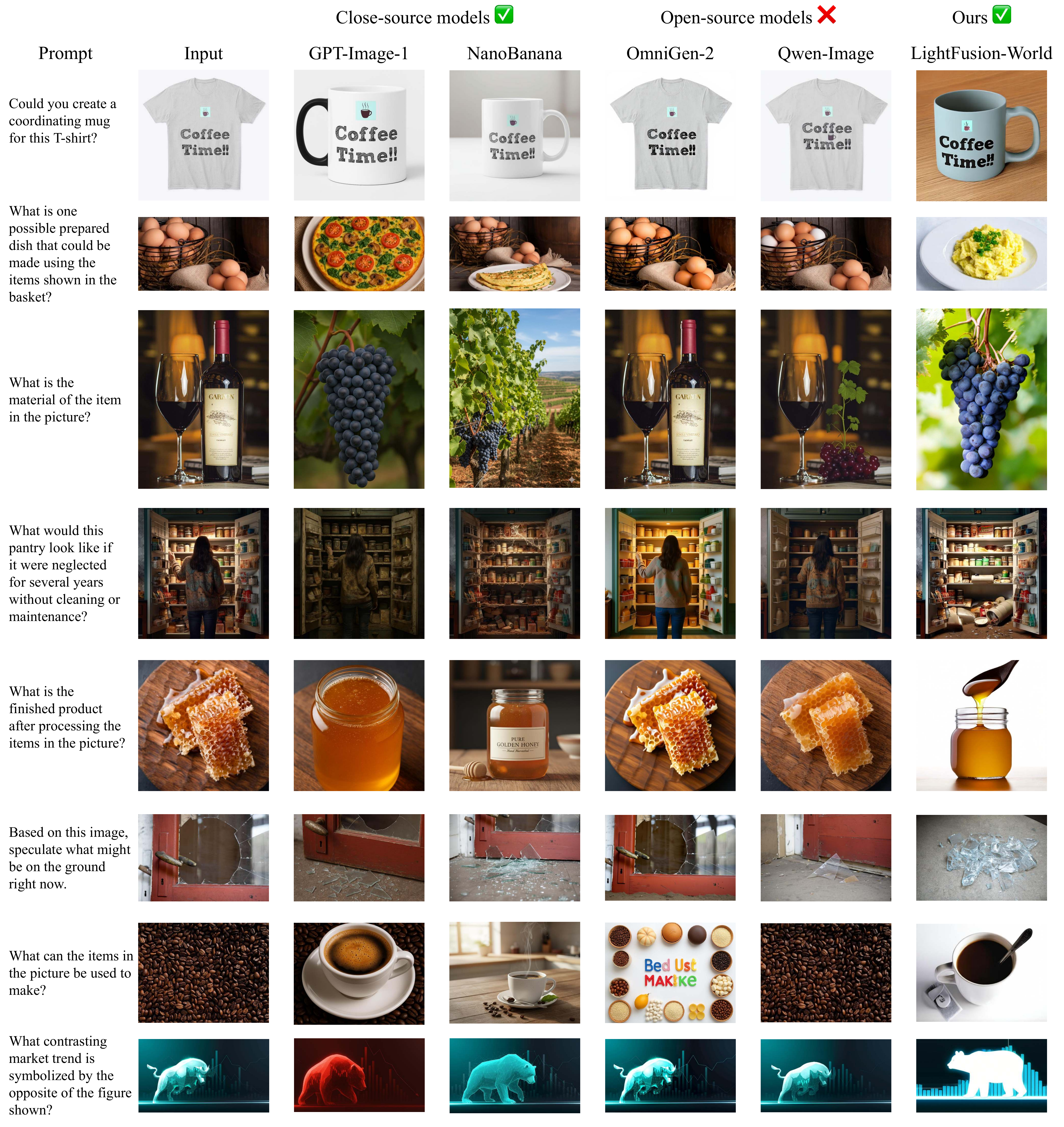}
    \vspace{-2em}
    \caption{Examples of Visual Question-Visual Answering (VQ-VA), highlighting the substantial gap between existing closed-source models and open-weight models. The rightmost column further shows that a model trained with \textsc{VQ-VA World} dataset significantly improves its VQ-VA performance.
    }\label{fig:1}
    \vspace{-.5em}
\end{figure*}

Driven by rapid advances in large multimodal generative models, frontier systems such as GPT-Image \citep{openai_gpt_image1} and NanoBanana \citep{nanobananaai} now demonstrate exceptionally strong image generation and editing capabilities, showing reliable instruction following, high-fidelity synthesis, and improved consistency. Beyond these strengths, they also begin to exhibit an emergent ability we term \textit{Visual Question-Visual Answering (VQ-VA)}, \emph{i.e.}, responding to a visual question with an image. As illustrated in Figure~\ref{fig:1}, when given a photo of a broken window and asked to speculate about what might be on the ground, NanoBanana generates an image depicting shards of glass; when shown an illustration of the stock market with a bull and asked "What is the contrasting trend?", NanoBanana creates an image of a bear to represent a bearish market. Producing such visual answers requires conditioning on the input image and instruction, and, more critically, leveraging internalized world knowledge and multi-step reasoning to yield contextually coherent outputs.

Despite this progress, VQ-VA remains largely restricted to proprietary systems such as GPT-Image and NanoBanana. As evident in Figure~\ref{fig:1}, current open-source models consistently underperform on these tasks: they often misinterpret the question or lack the world knowledge needed to synthesize an appropriate visual answer. We hypothesize that the primary bottleneck is data scarcity---open-source solutions are predominantly trained on standard image-editing datasets that emphasize predefined operations (\textit{e.g.}, object addition, removal, replacement, style transfer), while underrepresenting free-form visual generation that demands knowledge and multi-step reasoning.

In this paper, we present \textsc{VQ-VA World}, a data-driven framework to bridge this gap. At its core is an agentic data-construction pipeline with five modules: (1) Retriever---identifies semantically and knowledge-driven image pairs from web-interleaved documents; (2) Instruction Generator---produces free-form questions that require knowledge and reasoning, conditioned on the first image and using the second image as the answer; (3) Filter---automatically removes low-quality questions or pairs; (4) Rewriter---rephrases questions to enhance linguistic diversity; and (5) Reasoner---generates a natural-language reasoning trace that explains how to approach the question, what knowledge is required, and the detailed transformation from the source image to the target image.

\begin{table*}[t]
\centering
\caption{Comparison of major image-to-image datasets. QA indicates whether the dataset's instructions are formatted as questions rather than direct prompts. Knowledge-centric denotes whether the instructions require world knowledge. Real image is marked true only if both the input and output images are real for the majority of the dataset. Concepts refers to the number of distinct words appearing in the instructions. Note: For SEED-Data-Edit, only a small subset (0.073M out of 3.7M) contains real images.}\label{tab:edit-datasets}
\vspace{-0.5em}
\resizebox{0.7\linewidth}{!}{\begin{tabular}{lcccccc}
\toprule
\multirow{2}{*}{\shortstack{\textbf{Dataset}\\\textbf{(image-to-image)}}} 
 & \multirow{2}{*}{\textbf{\#Size}} & \multirow{2}{*}{\textbf{\shortstack{Freeform}}} & \multirow{2}{*}{\textbf{\shortstack{QA}}} & \multirow{2}{*}{\textbf{\shortstack{Knowledge\\Centric}}} & \multirow{2}{*}{\textbf{\shortstack{Real\\Image}}} & \multirow{2}{*}{\textbf{Concepts}} \\
 & & & & & & \\
\midrule
MagicBrush \citep{magicbrush}         & 10K   &\xmark  &\xmark  &\xmark  & \cmark  & 2K    \\
InstructPix2Pix \citep{instructpix2pix}    & 313K  &\xmark  &\xmark  &\xmark  &\xmark   & 11.6K \\
HQ-Edit \citep{hqedit}           & 197K  &\xmark  &\xmark  &\xmark  &\xmark   & 3.7K  \\
SEED-Data-Edit \citep{seed-x-edit}     & 3.7M  &\xmark  &\xmark  &\xmark  &\xmark   & 29.2K \\
UltraEdit \citep{ultraedit}         & 4M    &\xmark  &\xmark  &\xmark  &\xmark   & 3.7K  \\
AnyEdit \citep{anyedit}           & 2.5M  &\xmark  &\xmark  &\xmark  &\xmark   & 6.4K  \\
ImgEdit \citep{imgedit}           & 1.2M  &\xmark  &\xmark  &\xmark  &\xmark   & -     \\
MetaQuery \citep{metaquery}         & 2.4M  & \cmark &\xmark  &\xmark  & \cmark  & -     \\
\midrule
\textbf{Ours}      & 1.8M    & \cmark & \cmark & \cmark & \cmark  & 87.9K \\
\bottomrule
\end{tabular}}
\vspace{-1.5em}
\end{table*}

Deployed at web scale, this pipeline successfully curates 1.8M high-quality, interleaved image-text training samples across three subdomains: world knowledge (covering scientific, spatial, temporal, and other real-world domains), design knowledge, and reasoning. Moreover, to systematically assess models' VQ-VA capability, we introduce IntelligentBench, a human-curated benchmark sourced from real-world, web-interleaved documents. Each item is designed to probe specific knowledge and reasoning demands in VQ-VA. 
Additionally, we leverage leading VLMs (\emph{e.g.}, GPT-4o \citep{gpt4o} and Gemini-2.5-Flash \citep{gemini25}) as automatic judges to facilitate large-scale evaluation.

To evaluate the effectiveness of the \textsc{VQ-VA World} dataset, we fine-tune LightFusion \citep{LightFusion} (a fully open-source model; details provided in the Supp. files) on the 1.8M curated training samples and evaluate it on IntelligentBench. The results are striking: while previous open-source models achieve only trivial performance (\emph{e.g.}, 7.78 for LightFusion and 1.94 for UniWorld-V1), our LightFusion-World lifts the performance to 53.06, as shown in Table~\ref{tab:intelligent}. Similar improvements are also observed on other VQ-VA-related benchmarks such as RISEBench \citep{risebench} and KRIS-Bench \citep{krisbench} (see Table~\ref{tab:rise_kris}). More excitingly, our model surpasses several large models pretrained on massive private data across IntelligentBench and other VQ-VA-related benchmarks; for example, it outperforms Qwen-Image \citep{qwen-image} and FLUX.1-Kontext-Dev \citep{flux-kontext} on IntelligentBench, and surpasses Gemini-2.0-Flash \citep{gemini2}, Seedream-4.0 \citep{seedream4}, and BAGELThink \citep{bagel} on RISEBench \citep{risebench}. In addition, our results substantially narrow the gap with leading proprietary systems such as NanoBanana \citep{nanobananaai} and GPT-Image \citep{openai_gpt_image1}, as summarized in Tables~\ref{tab:intelligent} and \ref{tab:rise_kris}.

With the full release of model checkpoints, training and evaluation sets, and pipelines, we believe this work can help accelerate and inspire future open research in Visual Question-Visual Answering.

\section{Related Work}
\noindent\textbf{Image-to-Image models.} Existing Image-to-Image (I2I) models can be broadly categorized into three types: (1) single I2I models, (2) unified multimodal models for both understanding and generation, and (3) leading proprietary models. For single I2I models, InstructPix2Pix \citep{instructpix2pix} leverages synthetic data generated by GPT-3 \citep{gpt3} and Stable Diffusion \citep{SD} to train a conditional diffusion model capable of following human-written editing instructions. Emu Edit \citep{emuedit} is also diffusion-based, but it is trained on a diverse spectrum of editing tasks, including region-based I2I, free-form editing, and traditional computer vision tasks. Modern single I2I models such as Step1X-Edit \citep{step1xedit}, FLUX.1-Kontext \citep{flux-kontext}, and Qwen-Image \citep{qwen-image} have substantially improved editing performance through both data scaling and model scaling. In parallel, unified multimodal models \citep{chameleon, transfusion, metaquery, bagel, uniworld, blip3o} have gained popularity, benefiting from strong performance and cross-task learning advantages by combining understanding and generation. As for proprietary models, NanoBanana \citep{nanobananaai} and GPT-Image \citep{openai_gpt_image1} still exhibit a noticeable advantage over all other models, particularly showing emerging abilities on I2I tasks that require world knowledge and reasoning. The main motivation of our work is to narrow this gap in this specific domain for the open-source community.

\noindent\textbf{Public I2I datasets.} 
MagicBrush \citep{magicbrush} introduces a manually annotated dataset containing 10k triplets, covering four types: single-turn, multi-turn, mask-provided, and mask-free editing. HQ-Edit \citep{hqedit} builds a scalable data collection pipeline leveraging GPT-4V \citep{gpt4} and DALL-E 3 \citep{dalle3}, resulting in around 200k editing samples. UltraEdit \citep{ultraedit} employs an automatic pipeline that integrates an LLM and SDXL \citep{sdxl}, presenting a 4M-scale dataset consisting of real input images and synthetic edited images. SEED-Data-Edit \citep{seed-x-edit} proposes a hybrid dataset constructed from both human annotation and automatic pipelines, and further introduces specifically designed high-quality multi-turn image-editing data. OmniEdit-1.2M \citep{omniedit} is built using seven different specialist models and employs an importance sampling strategy to improve data quality. ImgEdit \citep{imgedit} and AnyEdit2.5 \citep{anyedit} expand the coverage of editing types to 13 and 25, respectively, thereby enhancing the instruction diversity of image-editing datasets. More recently, motivated by the strong performance of GPT-Image \citep{openai_gpt_image1} in generation tasks, GPT-IMAGE-EDIT-1.5M \citep{gpt-image-edit} relabels previous OmniEdit, HQ-Edit, and UltraEdit datasets using GPT-Image API, further improving the quality of open-source image-editing resources. 

Despite their scale and variety, these exisiting datasets are purpose-built for standard pixel-level editing: the target image is a direct modification of the source, guided by an explicit instruction. They thus under-represent scenarios that demand external knowledge and multi-step reasoning. Our \textsc{VQ-VA World} corpus instead targets VQ-VA, where the model must synthesize an entirely new image by leveraging real-world knowledge and reasoning, not merely edit the original.

\noindent\textbf{I2I benchmarks.}
EmuEdit Benchmark \citep{emuedit} covers 7 fixed editing types and adopts L1, CLIP-I, and DINO as scoring metrics to evaluate editing ability. MagicBrushEdit Benchmark \citep{magicbrush} extends this to 9 predefined tasks and provides two modes: mask-free and mask-provided. ImageEdit \citep{imgedit} further expands to 14 tasks, introduces VLM-based scoring, and supports multi-turn editing with varying difficulty levels. OMNI-EDIT-Bench \citep{omniedit} is a high-resolution, multi-aspect-ratio, multi-task benchmark comprising 434 edits derived from 62 images, evaluated with both VLM scorers and human judgments. GEdit-Bench \citep{step1xedit} contains 606 real-world user editing cases, filtered by humans and scored with VLMs. All of these datasets focus on standard image editing, whereas our work addresses VQ-VA, where the model must synthesize an entirely new image by leveraging knowledge and reasoning.

Two more recent benchmarks move closer to this setting: RISEBench \citep{risebench} and KRIS-Bench \citep{krisbench} emphasize reasoning and world knowledge, and several of their examples can be cast as VQ-VA. Our evaluation set, IntelligentBench, however, differs in two key respects:
(1) RISEBench and KRIS-Bench still primarily reward accurate pixel-level edits, while IntelligentBench deliberately includes tasks that require high-level semantic reasoning beyond what is visible in the source image (see Fig. \ref{fig:1}); and (2) both RISEBench and KRIS-Bench rely heavily on synthetic images, whereas IntelligentBench is curated from real-world web content; every item is manually verified and paired with a genuine reference answer image.

\section{Methods}
This section elaborates on the details of the \textsc{VQ-VA World} data framework and IntelligentBench.

\subsection{VQ-VA World Data Framework}
\textbf{Motivation.} The \textsc{VQ-VA World} framework tackles two key challenges: 1) identifying suitable data for VQ-VA and 2) designing a scalable pipeline for its construction. We target image pairs whose transformations (Image1 $\Leftrightarrow$ Image2) inherently require knowledge or reasoning---for example, (car wheel $\Leftrightarrow$ car), (mathematical equation $\Leftrightarrow$ its graph), or (window of a house $\Leftrightarrow$ broken glass on the ground). Such transformations capture semantic-level connections rather than superficial pixel-level alterations. By providing an image and formulating transformation-related questions whose answers require generating their corresponding counterparts, models can be trained to acquire knowledge-related VQ-VA ability. The subsequent step is to identify data sources rich in such pairs and to develop automated pipelines for large-scale collection and refinement. Inspired by the data used in LLM pretraining, we regard web-interleaved documents as a particularly promising candidate, since they naturally contain extensive world knowledge alongside closely associated images and text. Our target is to develop a pipeline that mines these image-text interleaved web documents and converts them into high-quality VQ-VA training triples.

\begin{figure*}
    \centering
    \includegraphics[width=0.95\linewidth]{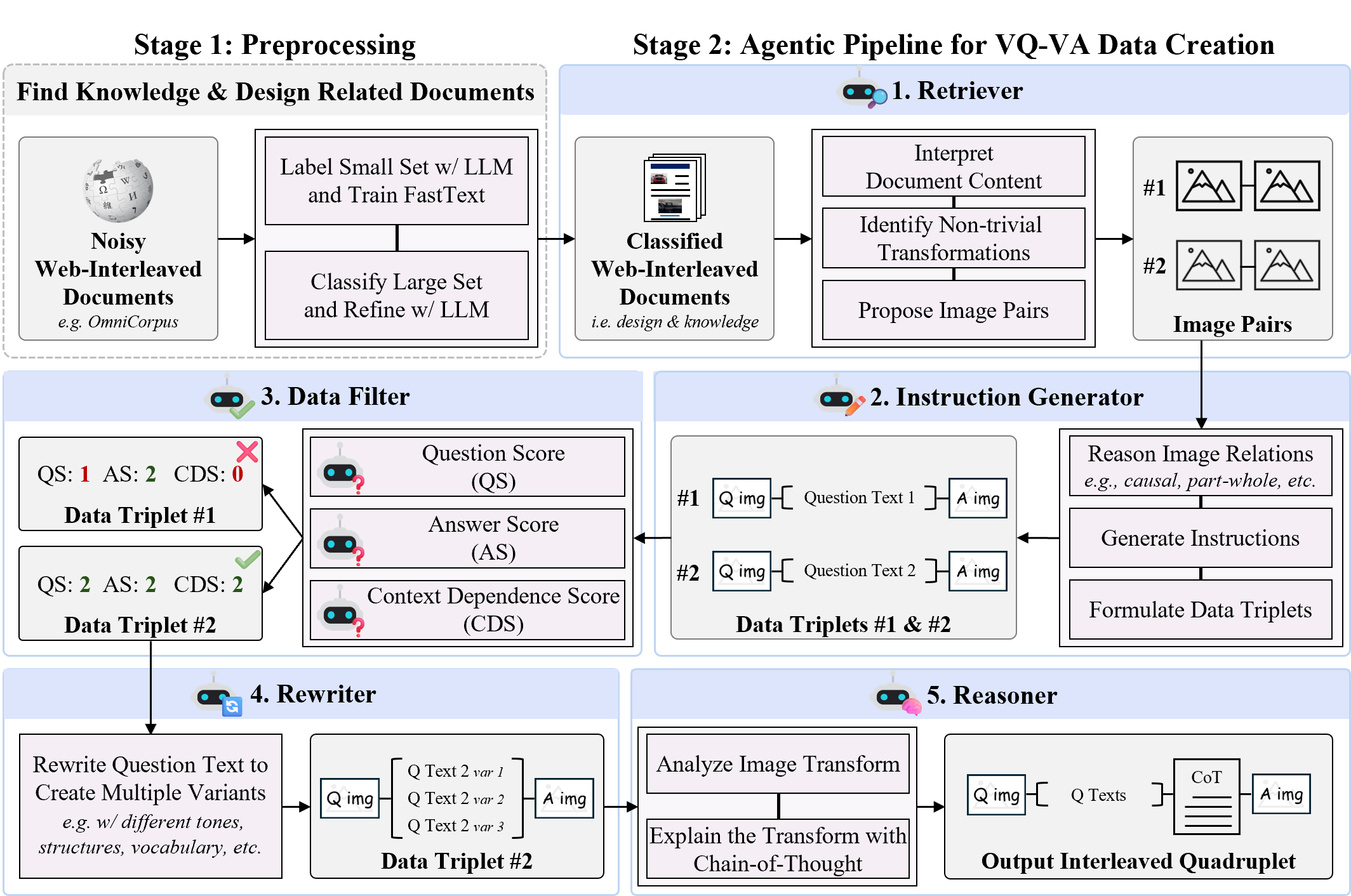}
    \vspace{-0.5em}
\caption{Illustration of the \textsc{VQ-VA World} framework for creating VQ-VA data. The framework consists of two stages: (1) preprocessing, which classifies and filters web-interleaved documents, and (2) an agentic pipeline that generates VQ-VA samples from the filtered documents. The agentic pipeline contains five sub-modules: retriever, filter, instruction generator, rewriter, and reasoner.}
\vspace{-1em}
\label{fig:pipeline}
\end{figure*}

\noindent\textbf{Framework Overview.}
As illustrated in Fig.~\ref{fig:pipeline}, \textsc{VQ-VA World} operates in two stages: data preprocessing and an agentic pipeline for VQ-VA data construction. In the preprocessing stage, noisy web-interleaved documents are processed and assigned semantic labels, with only those belonging to the knowledge and design categories retained. The agentic pipeline then transforms the filtered documents into high-quality VQ-VA samples. Running this pipeline at web scale produces a large-scale, high-quality training dataset with $\sim$1.8M samples, comprising 24.35\% reasoning, 30.37\% design knowledge, and 43.69\% world knowledge. We details each step below.

\noindent\textbf{Step 1: Preprocessing.}
The first challenge is to sift through web-scale corpora and isolate documents whose images are tied together by substantive, knowledge-rich relationships. We leverage a common prior that images on a webpage revolve around the page's central topic, making topic classification an effective proxy for relevance. Since the topic is not directly provided in web data, we design a loop to label documents efficiently, inspired by the data pipeline proposed in DeepSeek-Math \citep{deepseekmath}. Specifically, we first prompt an LLM (\emph{e.g.}, Qwen2.5-14B \citep{qwen25} in our case) to label a subset of the data and identify samples of the required types. The labeled data are then used to train a lightweight FastText \citep{fasttext} classifier, which enables large-scale labeling with high efficiency. Lastly, we apply an LLM again to refine the coarse labels produced by FastText. The final outputs of preprocessing are web-interleaved documents containing knowledge- and design-related content. The web document sources were collected from publicly available data \citep{li2025omnicorpus} in compliance with copyright and GDPR guidelines.

\label{sec: method}

\noindent\textbf{Step 2: Agent Pipeline for VQ-VA Data Creation.}
Our second stage turns the pre-filtered web-interleaved documents into high-quality VQ-VA examples. To scale the process and keep it modular, we design an "agentic" pipeline in which five independent workers handle a specific sub-task.

Specifically, each worker is powered by advanced VLMs (\emph{e.g.}, GPT-4o \citep{gpt4o} and Seed1.5VL-Thinking \citep{seed1_5vl}), and is guided by carefully designed system prompts and chain-of-thought reasoning, without memory sharing across workers. We define the agent workers below:

(1) \textit{Agent Retriever} selects image pairs from interleaved documents that can serve as the basis for free-form questions. It focuses on pairs with meaningful transformations, especially those involving non-trivial relations grounded in knowledge and reasoning. We also find it beneficial for the retriever to capture the document's topic; hence, its input is the full document rather than merely the image list. The detail prompt is provided in Supp. Table~\ref{tab:prompt_retriver}.

(2) \textit{Agent Instruction Generator} write a natural-language question about one image so that the other image serves as the correct answer. For instance, for the pair (car wheel $\Leftrightarrow$ racing car), if the question image is the wheel, it might ask: "What is it used for?" The questions are designed to probe diverse forms of knowledge and reasoning, including but not limited to: temporal or causal relations (\emph{e.g.}, an object before \emph{vs.} after an event, or sequential steps with clear causality); compositional or spatial structures (\emph{e.g.}, part-whole links, inside-outside contrasts, exploded or sectional views); and scientific or analytical phenomena (\emph{e.g.}, visual explanations of scientific or mathematical concepts). The detailed prompt is provided in Supp. Table~\ref{tab:prompt_question}.

(3) \textit{Agent Filter} removes low-quality triplets $\langle$Question Image, Question Text, Answer Image$\rangle$. Specifically, through careful multi-round human-in-the-loop audits, we identify several common issues leading to low-quality data, such as poorly formulated questions, ambiguous or irrelevant answer images, and context shortcuts (\emph{i.e.}, cases where the answer can be inferred from the text alone, making the question image unnecessary). To effectively address these issues, we design a multi-score VLM-based filtering strategy with three sub-scorers: Question Score (QS), Answer Score (AS), and Context Dependence Score (CDS). The detailed prompts are provided in Supp. Table~\ref{tab:prompt_question_filter}, ~\ref{tab:prompt_answer_filter} and ~\ref{tab:prompt_overall_mark}, respectively. Each score is assigned on a three-level scale ${0,1,2}$, and only cases with the maximum total (\emph{i.e.}, QS + AS + CDS = 6) are retained. In addition, we manually design and iteratively refine the scoring template, and adopt a chain-of-thought approach during scoring, where the model generates an analysis before assigning scores, thereby further enhancing filtering effectiveness.

(4) \textit{Agent Rewriter} increases instruction diversity by producing multiple variants of the original questions. The variants differ in tone, sentence structure, vocabulary, expression, and overall linguistic naturalness. This rewriting process is essential for improving instruction-following ability. The detail prompt is provided in Supp. Table~\ref{tab:prompt_rewrite}.

(5) \textit{Agent Reasoner} generates a language-based chain-of-thought explanation describing how the source image should be transformed to obtain the target image. The process involves analyzing the question, observing the question image, identifying changes, determining which elements remain consistent, and highlighting key modifications. This reasoning trace is then incorporated with the triplet to construct a new data-format quadruplet $\langle$Question Image, Question Text, Editing reasoning trace, Answer Image$\rangle$. This quadruplet is used to fine-tune a unified multimodal model, \emph{i.e.}, LightFusion, to improve both reasoning-trace generation and instruction-following ability. The detailed prompt is provided in Supp. Table~\ref{tab:prompt_reasoner}.
\\ 
 \begin{figure}
     \begin{center}
     \includegraphics[width=1.0\linewidth]{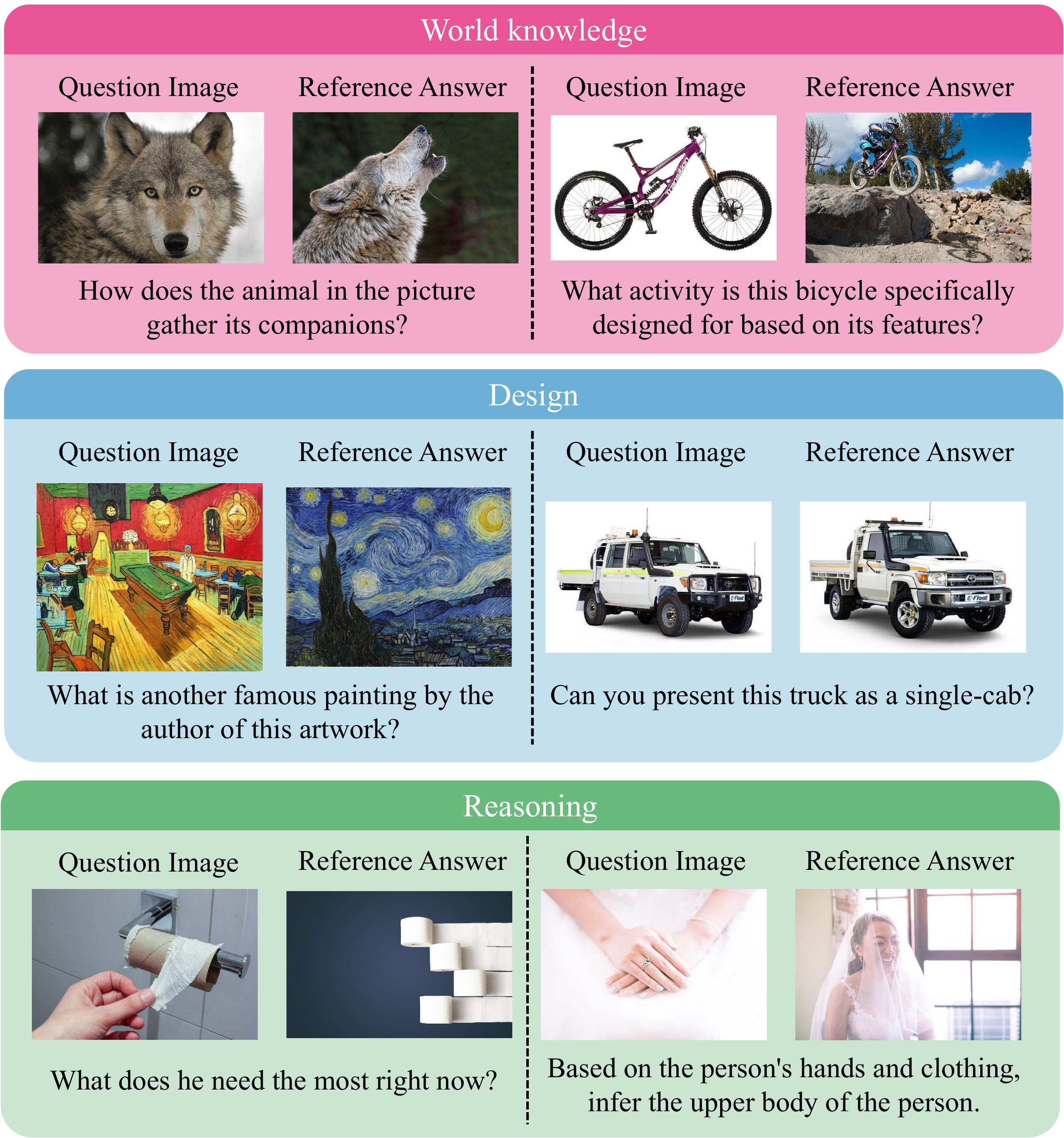} 
     \vspace{-1.5em}
\caption{Illustration of the three question types in IntelligentBench. Each type is shown with two examples, and each example contains a question image, question text, and the answer image.}
   \end{center}
  \vspace{-2.5em}
 \end{figure}
\\\noindent\textbf{High-quality subset curation.} Following prior works such as \citep{bagel,qwen-image}, which typically adopt multi-stage training, we employ a two-stage strategy: continued pretraining and supervised fine-tuning (SFT). In the first stage, we train on the full large-scale dataset for additional steps to strengthen knowledge and instruction-following ability. In the second stage, we focus on a smaller high-quality subset for fewer steps to improve quality. Specifically: (1) we apply stricter filtering, retaining the best one-third of the data, which yields about 500k high-quality samples; and (2) leveraging the fact that video models naturally encode temporal knowledge, we use the Seedance video model \citep{gao2025seedance} to construct a set of $\sim$100k temporally related VQ-VA samples.

\subsection{IntelligentBench}
\textbf{Benchmark data.} The purpose of IntelligentBench is to evaluate the VQ-VA abilities of different models, where the questions require knowledge and reasoning to answer. 
Specifically, it contains 360 human-curated examples divided into three domains---world knowledge (171), design knowledge (88), and reasoning (101). 
The construction of IntelligentBench involves three main steps: (1) Document Review: Human experts examined about 3k classified interleaved web documents and, from each, selected the image pair that best represented the document's content and exhibited strong semantic connections. (2) Question Design: For each selected image pair, experts designed free-form questions targeting world knowledge, design knowledge, or reasoning. (3) Expert Cross-Review: Each candidate item is independently reviewed by at least one additional expert; only items that receive unanimous approval are retained, resulting in 360 final examples.

\begin{figure}
    \centering
    \includegraphics[width=0.85\linewidth]{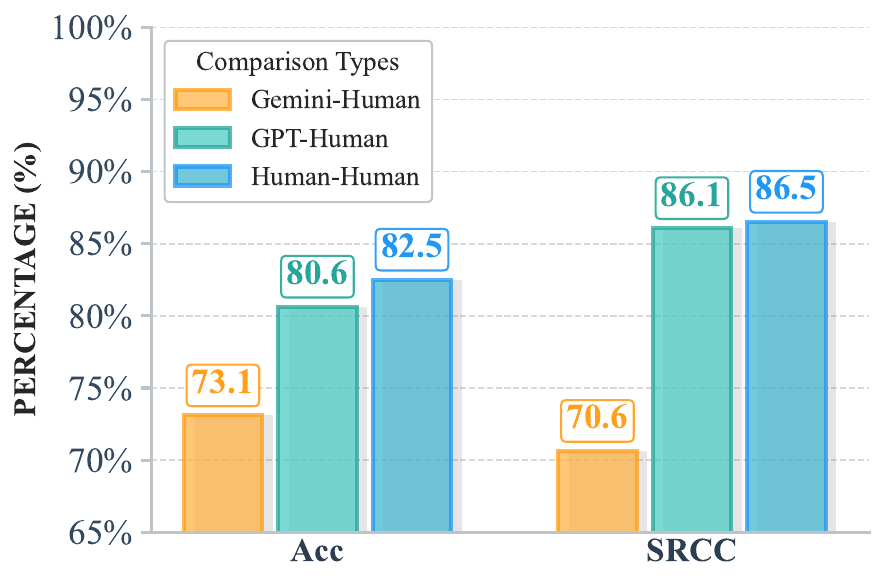} 
    \vspace{-1em}
    \caption{Alignment between VLM and human scores. We compare Gemini-2.5-Flash vs. human experts, GPT-4o vs. human experts, and agreement among human experts. We report the Accuracy and Spearman Rank Correlation Coefficient (SRCC) for comprehensive comparison.
}
    
    \label{fig:4:vlm agreement}
   \vspace{-1em}
\end{figure}

\noindent\textbf{Evaluation Metric.}
We use a VLM as the automatic judge, following rules: (1) the VLM is provided with the question image, question text, reference answer image, the generated image, and a carefully designed system prompt; (2) the VLM is required to output a score as an integer in \{0, 1, 2\}. The full rubric and prompt is provided in the Supp.

\begin{table*}[tbp]
\centering
\setlength{\fboxsep}{0pt}

\caption{Results on IntelligentBench, a benchmark designed for VQ-VA. 
\protect\Stars{1} refers to closed-source models. 
\protect\Stars{2} refers to open-weight models; \protect\Stars{3} refers to the fully open-source models (both full training data and model weights).
}

\label{tab:intelligent}

\vspace{-0.5em}
\begin{adjustbox}{width=0.75\linewidth,center}
\begin{tabular}{l c c c c c}
\toprule
\multirow{2}{*}{\textbf{Model}} & 
\multirow{2}{*}{\textbf{Open Source Level}} & 
\multirow{2}{*}{\textbf{\shortstack{World\\Knowledge}}} & 
\multirow{2}{*}{\textbf{\shortstack{Design\\Knowledge}}} & 
\multirow{2}{*}{\textbf{Reasoning}} & 
\multirow{2}{*}{\textbf{Overall}} \\
 & & & & \\
\midrule
GPT-Image-1 \citep{openai_gpt_image1}  & \Stars{1} & 84.5  & 80.68 & 81.19 & 82.64 \\
Nano Banana \citep{nanobananaai}   &\Stars{1} & 81.6  & 82.95 & 80.69 & 81.67 \\
BAGELThink \citep{bagel}   &\Stars{2} &61.99 & 55.11 & 62.38 & 60.42 \\
Qwen-Image \citep{qwen-image} &\Stars{2}    & 38.07 & 33.66 & 32.75 & 34.31 \\
FLUX.1-Kontext-Dev \citep{flux-kontext} &\Stars{2} &   20.18 & 24.43 &  19.80 & 21.11 \\
OmniGen2 \citep{omnigen2} &\Stars{2}  & 11.11 & 13.07 & 7.92  & 10.69 \\
Step1X-Edit \citep{step1xedit}  &\Stars{2}   & 11.7  & 10.23 & 15.35 & 12.36 \\
\midrule
UniWorld-V1 \citep{uniworld} & \Stars{3} & 2.92    &0.57       &  1.49     & 1.94      \\
\textbf{LightFusion} \citep{LightFusion}  &\Stars{3} & 5.26       & 11.93     & 8.42      & 7.78  \\
\textbf{LightFusion-World} & \Stars{3} & \textbf{50.58} & \textbf{57.95} & \textbf{52.97} & \textbf{53.06} \\
\bottomrule

\end{tabular}
\end{adjustbox}
\vspace{-0.5em}
\end{table*}

\noindent\textbf{Metric Validation.} 
To validate the reliability of our automatic grading process, we conducted a comparative evaluation involving four human experts and two state-of-the-art VLMs, each independently scoring outputs from four different models. Human inter-annotator agreement averaged 82.5\%. As illustrated in the left panel of Figure~\ref{fig:4:vlm agreement}, GPT-4o~\citep{gpt4o} achieved 80.6\% agreement with human ratings, while Gemini-2.5-Flash~\citep{gemini25} achieved 73.1\%. The Spearman Rank Correlation Coefficient (SRCC) followed the same trend, indicating that GPT-4o's evaluations most closely reflect human judgment. We therefore adopt GPT-4o as the default evaluator for IntelligentBench.

\section{Experiments}
\label{sec: exp}

\setlength{\tabcolsep}{2pt}
\setlength{\tabcolsep}{3pt}
\begin{table*}[t]
\centering
\caption{
Combined results on two reasoning-centric image editing benchmarks, RISEBench and KRIS-Bench. For previously published models, we directly cite their official results reported in papers or public leaderboards. For LightFusion and our fine-tuned models, we follow their official evaluation pipeline to reproduce and report the corresponding test results. \protect\Stars{1} refers to closed-source models. 
\protect\Stars{2} refers to open-weight models; \protect\Stars{3} refers to the fully open-source models (both full training data and model weights).
}

\label{tab:rise_kris}
\begin{adjustbox}{width=0.95\linewidth,center}
\begin{tabular}{l c | c c c c c | c c c c}
\toprule
\multirow{2}{*}{\bf Model} &
\multirow{2}{*}{\bf \textbf{Open Source Level}} &
\multicolumn{5}{c|}{\bf RISEBench} &
\multicolumn{4}{c}{\bf KRIS-Bench} \\
\cmidrule(lr){3-7} \cmidrule(lr){8-11}
& & {\bf Temporal} & {\bf Causal} & {\bf Spatial} & {\bf Logical} & {\bf Overall} & 
{\bf Factual} & {\bf Conceptual} & {\bf Procedural} & {\bf Average} \\
\midrule


Nano Banana \citep{nanobananaai} & \Stars{1} 
& 25.9 & 47.8 & 37.0 & 18.8 & 32.8
& -- & -- & -- & -- \\

GPT-Image-1 \citep{gpt-image-edit} & \Stars{1}
& 34.1 & 32.2 & 37.0 & 10.6 & 28.9
& 79.80 & 81.37 & 78.32 & 80.09 \\

Gemini-2.0-Flash \citep{gemini2} & \Stars{1}
& 8.2 & 15.5 & 23.0 & 4.7 & 13.3
& 65.26 & 59.65 & 62.90 & 62.41 \\

Seedream-4.0 \citep{seedream4} & \Stars{1}
& 12.9 & 12.2 & 11.0 & 7.1 & 10.8
& -- & -- & -- & -- \\


BAGELThink \citep{bagel} & \Stars{2}
& 5.9 & 17.7 & 21.0 & 1.1 & 11.9
& 55.77 & 59.44 & 39.26 & 53.36 \\

Qwen-Image-Edit \citep{qwen-image} & \Stars{2}
& 4.7 & 10.0 & 17.0 & 2.4 & 8.9
& -- & -- & -- & -- \\

FLUX.1-Kontext-Dev \citep{flux-kontext} & \Stars{2}
& 2.3 & 5.5 & 13.0 & 1.2 & 5.8
& -- & -- & -- & -- \\

Step1X-Edit \citep{step1xedit} & \Stars{2}
& 0.0 & 2.2 & 2.0 & 3.5 & 1.9
& 45.52 & 48.01 & 31.82 & 43.29 \\

EMU2 \citep{emu2} & \Stars{2}
& 1.2 & 1.1 & 0.0 & 0.0 & 0.5
& 45.40 &  37.54 &  34.91 &  39.70 \\

HiDream-Edit \citep{hidream} & \Stars{2}
& 0.0 & 0.0 & 0.0 & 0.0 & 0.0
& -- & -- & -- & -- \\

FLUX.1-Canny \citep{flux-kontext} & \Stars{2}
& 0.0 & 0.0 & 0.0 & 0.0 & 0.0
& -- & -- & -- & -- \\

OmniGen \citep{omnigen1} & \Stars{2}
& 1.2 & 1.0 & 0.0 & 1.2 & 0.8
& 33.11 &  28.02 & 23.89 &  28.85 \\
\midrule
MagicBrush \citep{magicbrush} & \Stars{3}
& -- & -- & -- & -- & -- &
41.84 &  39.24 &  26.54 &  37.15 \\

AnyEdit \citep{anyedit} & \Stars{3}
& -- & -- & -- & -- & -- & 
39.26 &  41.88 &  31.74 &  38.55 \\
InsPix2Pix \citep{instructpix2pix} & \Stars{3}
& -- & -- & -- & -- & -- & 
 23.33 &   25.59 &   17.28 &   22.82 \\
\textbf{LightFusion} \citep{LightFusion}& \Stars{3}
& 2.4 & 4.4 & 9.0 & 0.0 & 4.2
& 60.44 & 51.23 & 44.83 & 52.52 \\

\textbf{LightFusion-World} & \Stars{3}
& \textbf{15.3} & \textbf{25.5} & \textbf{16.0} & \textbf{3.5} & \textbf{15.3}
& \textbf{66.69} & \textbf{63.50} & \textbf{52.38} & \textbf{61.85} \\
\bottomrule
  \vspace{-2.5em}
\end{tabular}
\end{adjustbox}
\end{table*}

\textbf{Implementation details.} We adopt the fully-open, light-training unified multimodal model, LightFusion~\citep{LightFusion}, as our baseline. 

Specifically, LightFusion leverages the publicly available Qwen2.5-VL-7B~\citep{qwen25} as the understanding branch and Wan2.2-TI2V-5B~\citep{wan2025} as the generation branch, and further introduces a double fusion approach to synergize these two branches. In our experiments, we incorporate \textsc{VQ-VA World} dataset into the overall training set of LightFusion with a sampling ratio of 25\%, and fine-tune the model for a total of 45k steps. Both branches are trained following LightFusion's default recipe with the timestep shift set to 4. We adopt a two-stage training scheme: (1) continued training of LightFusion with a mix of the 1.8M \textsc{VQ-VA World} dataset for 30k steps with AdamW and a cosine learning rate schedule (peak $1\times10^{-5}$). (2) supervised fine-tuning on a further filtered high-quality subset ($\sim$1/3 of the original \textsc{VQ-VA World} dataset) for 15k steps with a constant learning rate of $1\times10^{-5}$. Note that in both stages, the original 45M LightFusion data is mixed.

\noindent\textbf{Evaluation setting.} For a comprehensive evaluation of \textsc{VQ-VA World}, we consider three domains with five benchmarks:
(1) VQ-VA, evaluated on \textit{IntelligentBench};
(2) reasoning- and knowledge-informed image editing, evaluated on \textit{RISEBench} and \textit{KRIS-Bench}, with the results summarized in \cref{tab:rise_kris}; both benchmarks require pixel-level alignment and strong reasoning capability; and
(3) standard image editing, evaluated on \textit{GEdit-Bench} \citep{step1xedit}, constructed from real-world user editing cases, and \textit{ImgEdit-Bench} \citep{imgedit}, designed to assess instruction adherence, editing quality, and detail preservation.

Results on \textit{IntelligentBench} are shown in Table~\ref{tab:intelligent}; results on \textit{RISEBench} and \textit{KRIS-Bench} are shown in Table~\ref{tab:rise_kris}; and summarized results on traditional image editing tasks (\textit{GEdit-Bench} and \textit{ImgEdit-Bench}) are presented in Table~\ref{tab:trad_edit}. Following the setup in \citep{bagel}, for all knowledge-intensive benchmarks, the model is configured to first output reasoning content before generating the image, whereas for traditional image editing benchmarks, we directly generate the image. For all benchmarks, we adopt a double-CFG strategy when evaluating both our LightFusion-World and the baseline LightFusion, with the image CFG scale set to 2 and the text CFG scale set to 4. The time shift is fixed at 4 for both training and evaluation.

\begin{table}[t]
\centering
\caption{Results on Standard Image Editing Benchmarks (GEdit-Bench-EN and ImgEdit-Bench). Higher scores are better. \protect\Stars{1} refers to closed-source models. 
\protect\Stars{2} refers to open-weight models; \protect\Stars{3} refers to the fully open-source models (both full training data and model weights)
}

\label{tab:trad_edit}
\begin{adjustbox}{width=\linewidth,center}
\begin{tabular}{l c ccc c}
\toprule
\multirow{2}{*}{\textbf{Model}} & \multirow{2}{*}{\bf \textbf{Open Source Level}} &
\multicolumn{3}{c}{\textbf{GEdit-Bench-EN}} 
& \multicolumn{1}{c}{\textbf{ImgEdit-Bench}} \\
\cmidrule(lr){3-5} \cmidrule(lr){6-6}
& & SC & PQ & Overall & Overall \\
\midrule

GPT-4o \citep{gpt-image-edit} & \Stars{1}         & 7.85 & 7.62 & 7.53 & 4.20 \\
Gemini-2.0-Flash \citep{gemini2} & \Stars{1} & 6.73 & 6.61 & 6.32 & -    \\

ICEdit  \citep{icedit} & \Stars{2} & 5.11 & 6.85 & 4.84 & 3.05 \\
Step1X-Edit \citep{step1xedit} & \Stars{2}  & 7.09 & 6.76 & 6.70 & 3.06 \\

OmniGen2  \citep{omnigen2} & \Stars{2}  & 7.16 & 6.77 & 6.41 & 3.43 \\
BAGEL  \citep{bagel}  & \Stars{2}    & 7.36 & 6.83 & 6.52 & 3.20 \\
Ovis-U1  \citep{ovis} & \Stars{2}    & --   & --   & 6.42 & 3.98 \\
UniPic \citep{unipic} &  \Stars{2}   & 6.72 & 6.18 & 5.83 & 3.49 \\
UniPic 2.0  \citep{unipic2} & \Stars{2} & --   & --   & 7.10 & 4.06 \\
\midrule
Instruct-Pix2Pix \citep{instructpix2pix} & \Stars{3} & 3.58 & 5.49 & 3.68 & 1.88 \\
MagicBrush \citep{magicbrush} & \Stars{3}      & 4.68 & 5.66 & 4.52 & 1.90 \\
AnyEdit  \citep{anyedit}    & \Stars{3}     & 3.18 & 5.82 & 3.21 & 2.45 \\
UniWorld-V1 \citep{uniworld} & \Stars{3}          & 4.93 & \textbf{7.43} & 4.85 & 3.26 \\
\textbf{LightFusion} \citep{LightFusion}&\Stars{3}          & 6.34 & 7.31 & 6.06 & 3.77 \\
\textbf{LightFusion-World}  &\Stars{3}       & \textbf{7.00}	&7.29	& \textbf{6.58} & \textbf{3.85} \\
\bottomrule
\end{tabular}
\end{adjustbox}
\vspace{-1em}
\end{table}

\subsection{Results on VQ-VA}
We first evaluate LightFusion-World along with other advanced closed-source and open-source models on IntelligentBench. Scores are normalized to the range 0-100 for each domain and averaged across domains; items for which a model fails to produce an image receive a score of 0.

As reported in Table~\ref{tab:intelligent}, the results show that LightFusion-World achieves the best performance among fully open-source models, and the large gap between the baseline model LightFusion and LightFusion-World further supports the effectiveness of our dataset. Moreover, LightFusion-World even surpasses Qwen-Image, which was pretrained on large-scale proprietary data and adopted RL for further improvement. Lastly, when compared with leading proprietary models such as GPT-4o and Gemini, we can see that a performance gap remains but has already been substantially reduced. We provide more qualitative results of all models in Supp. Figure~\ref{fig:sub:design:1}-~\ref{fig:sub:knowledge:13}.

\subsection{Results on Reasoning-Based Image Editing Benchmark}
In this domain, we evaluate models on RISEBench and KRIS-Bench, as shown in Table~\ref{tab:rise_kris}. On RISEBench, the results indicate that: (1) our model achieves performance comparable to BAGEL-Think while requiring far less training data; (2) Relative to the vanilla LightFusion baseline, our model posts a large absolute gain; and (3) some large in-house-data-trained models such as Qwen-Image-Edit and FLUX.1-Kontext-Dev underperform ours, highlighting potential limitations of unbalanced data distribution and the necessity of free-form, knowledge-rich data like \textsc{VQ-VA World} dataset. KRIS-Bench exhibits the same pattern: LightFusion-World consistently outperforms every fully open-source competitor. These findings further support the effectiveness of \textsc{VQ-VA World} and the benefits brought by enhanced VQ-VA capability. More qualitative results on RISEBench are provided in Supp.~\ref{fig:rise_qualitative}.

\subsection{Results on Standard Image Editing Benchmark}
Lastly, we report standard image editing performance on GEdit-Bench-EN and ImgEdit-Bench, as shown in Table~\ref{tab:trad_edit}. The complete ImgEdit-Bench results for each subdomain (\textit{e.g.}, add/remove) are provided in the Supp. Table~\ref{tab:completeimgedit}. From these tables, we can see that our model delivers consistent gains over the LightFusion baseline on both datasets. This modest margin---especially when contrasted with the large improvements seen on VQ-VA and reasoning-centric editing---highlights the clear domain gap between routine pixel-level edits and knowledge-driven generation.

\subsection{Summarized Results} 
Combining \cref{tab:intelligent,tab:rise_kris,tab:trad_edit}, we make the following observations:
(1) Existing open-source models show certain ability on standard image editing, and the performance gap with closed-source models has been substantially reduced thanks to recent open image-editing dataset efforts. However, they still struggle on VQ-VA, and the gap remains significant. This further indicates the necessity of developing open-source VQ-VA-related data.
(2) With the help of VQ-VA data, LightFusion achieves clear improvements not only on VQ-VA but also on reasoning-based image editing tasks, along with noticeable gains on standard image editing. This supports the view that generalized VQ-VA capability also benefits other tasks.

\section{Conclusion}
This work focuses on studying VQ-VA, an emerging property that has already been \textit{exclusively} seen in leading proprietary models. To bring this capability to open-source models, we develop \textsc{VQ-VA World}, a scalable data-centric framework driven by an agentic pipeline for constructing high-quality, diverse VQ-VA training data.
Our web-scale pipeline curates $\sim$1.8 million high-quality samples, and we complemented it with IntelligentBench, a human-curated benchmark to rigorously assess the VQ-VA capability. Fine-tuning LightFusion on BAGEL-World lifts its IntelligentBench score from 7.78 to 53.06, surpassing all existing open-source models and substantially narrowing the gap to proprietary leaders. We are releasing the full suite of code, data, pipelines, and model checkpoints to spur further research on VQ-VA and, more broadly, on building more powerful multimodal systems that can \textit{answer with images}.

\newpage

{
    \small
    \bibliographystyle{ieeenat_fullname}
    \bibliography{main}
}

\newpage

\begin{figure*}[ht]
    \centering
    \includegraphics[width=\linewidth]{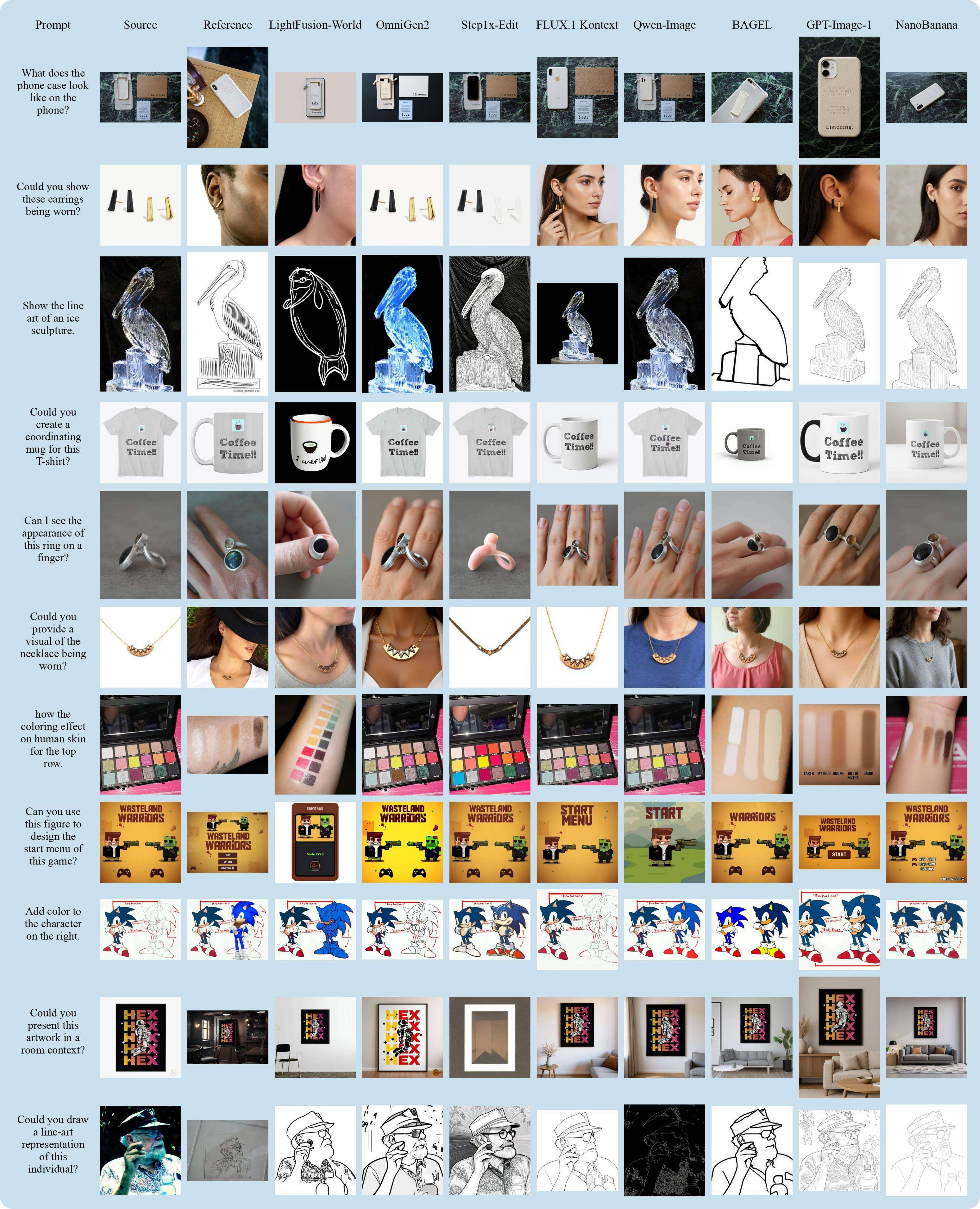}
    \caption{Comprehensive visualization of model performance on IntelligentBench (Subset Design, part 1/9).}\label{fig:sub:design:1}
\end{figure*}

\begin{figure*}[ht]
    \centering
    \includegraphics[width=\linewidth]{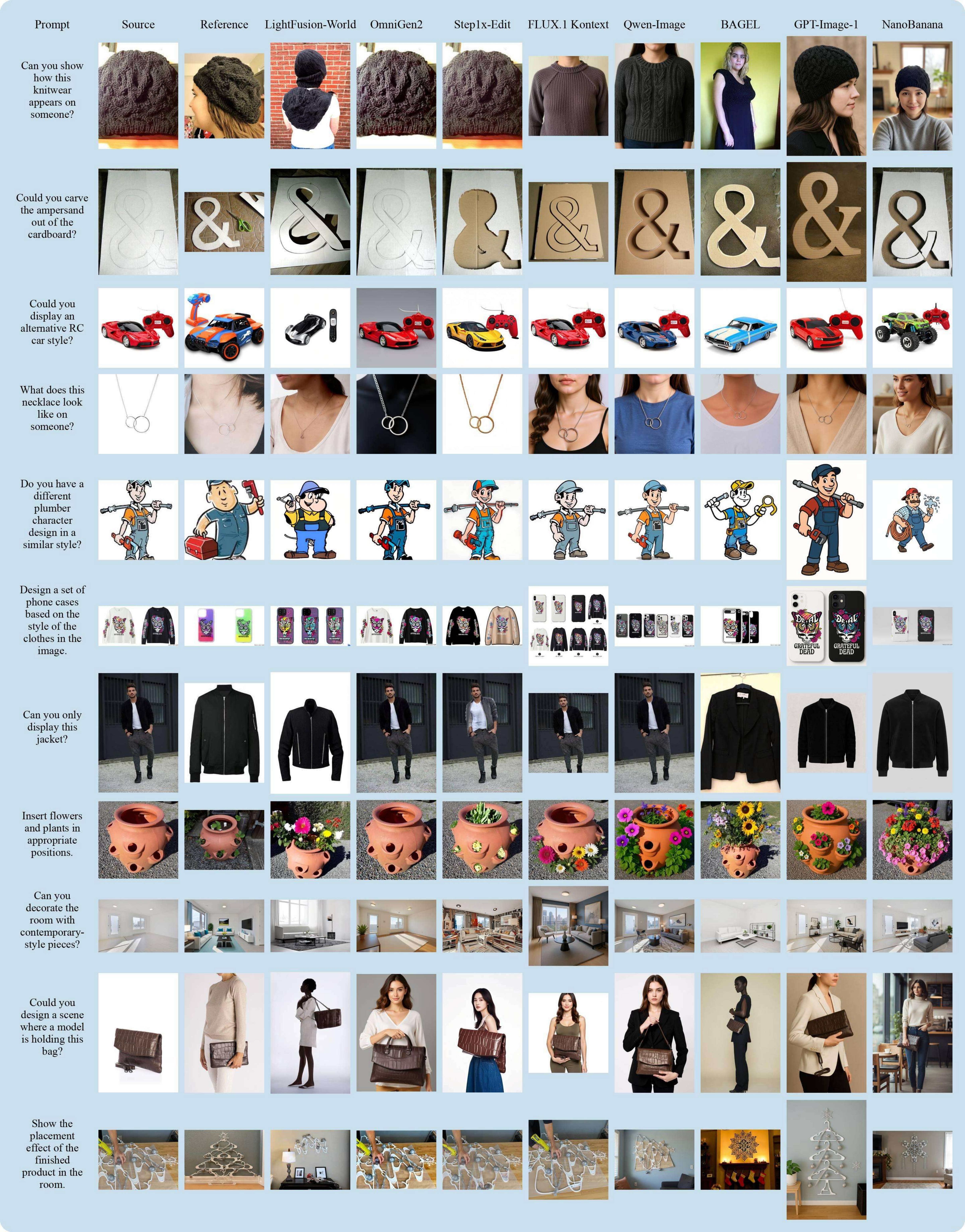}
    \caption{Comprehensive visualization of model performance on IntelligentBench (Subset Design, part 2/9).}\label{fig:sub:design:2}
\end{figure*}

\begin{figure*}[ht]
    \centering
    \includegraphics[width=\linewidth]{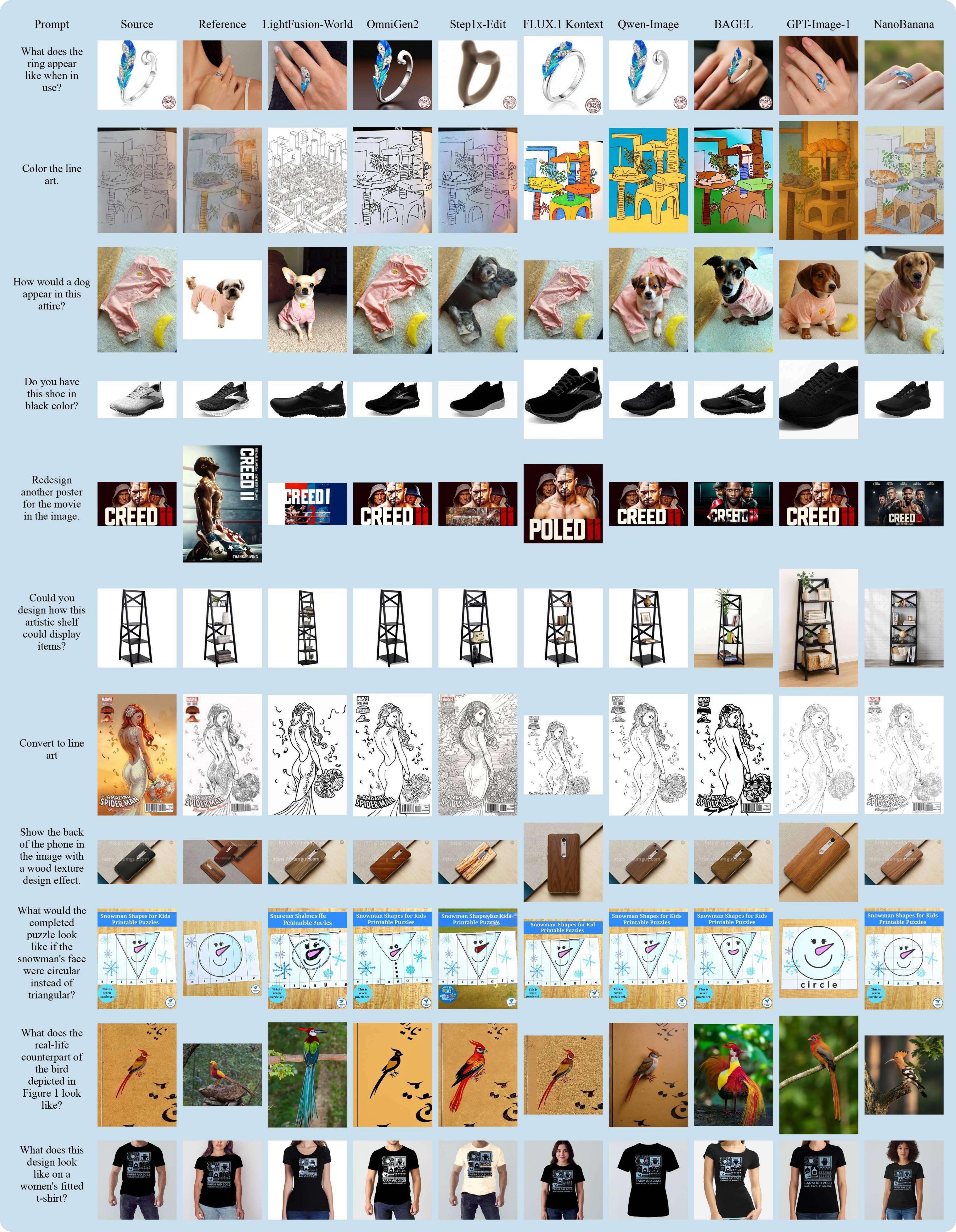}
    \caption{Comprehensive visualization of model performance on IntelligentBench (Subset Design, part 3/9).}\label{fig:sub:design:3}
\end{figure*}

\begin{figure*}[ht]
    \centering
    \includegraphics[width=\linewidth]{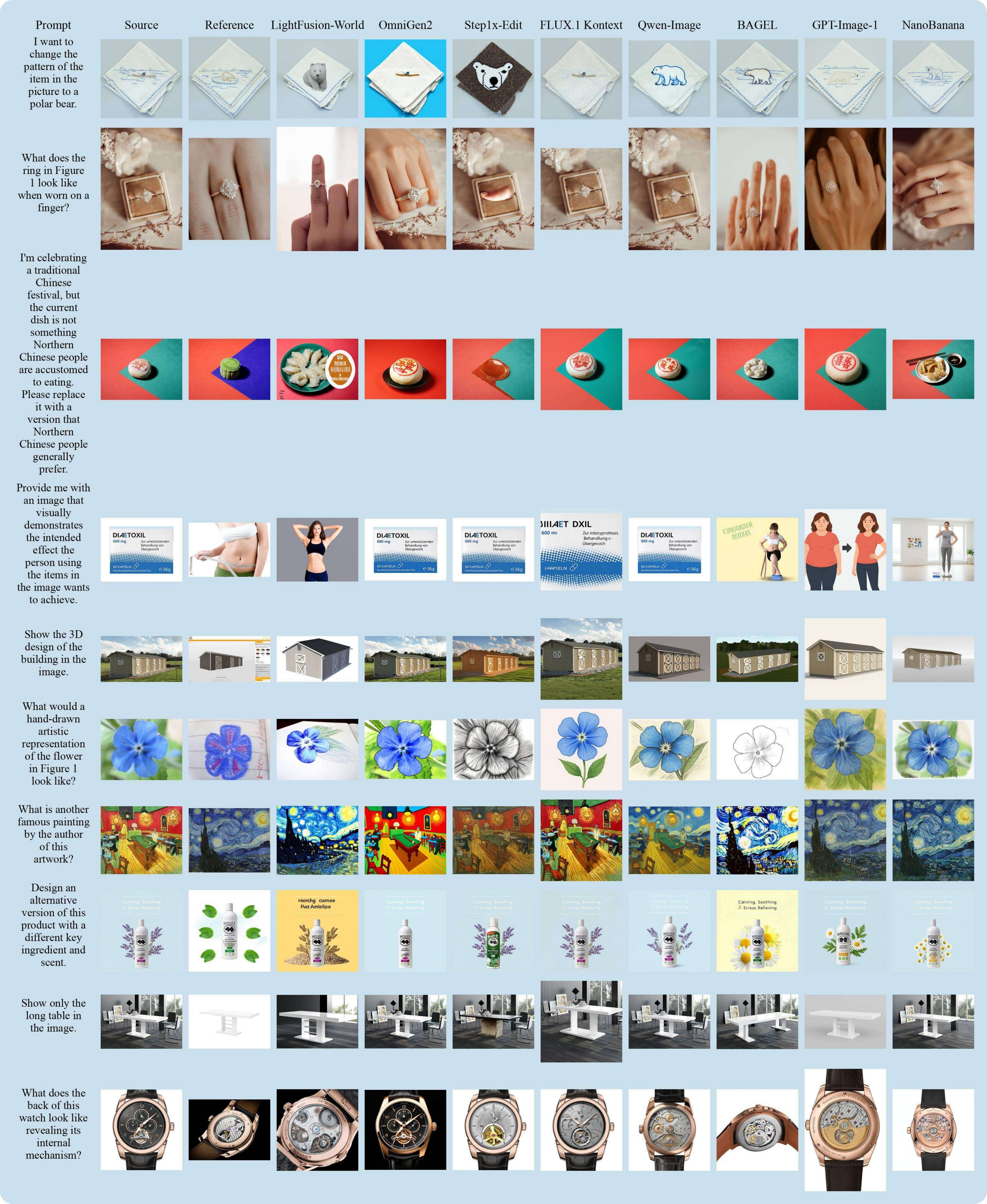}
    \caption{Comprehensive visualization of model performance on IntelligentBench (Subset Design, part 4/9).}\label{fig:sub:design:4}
\end{figure*}

\begin{figure*}[ht]
    \centering
    \includegraphics[width=\linewidth]{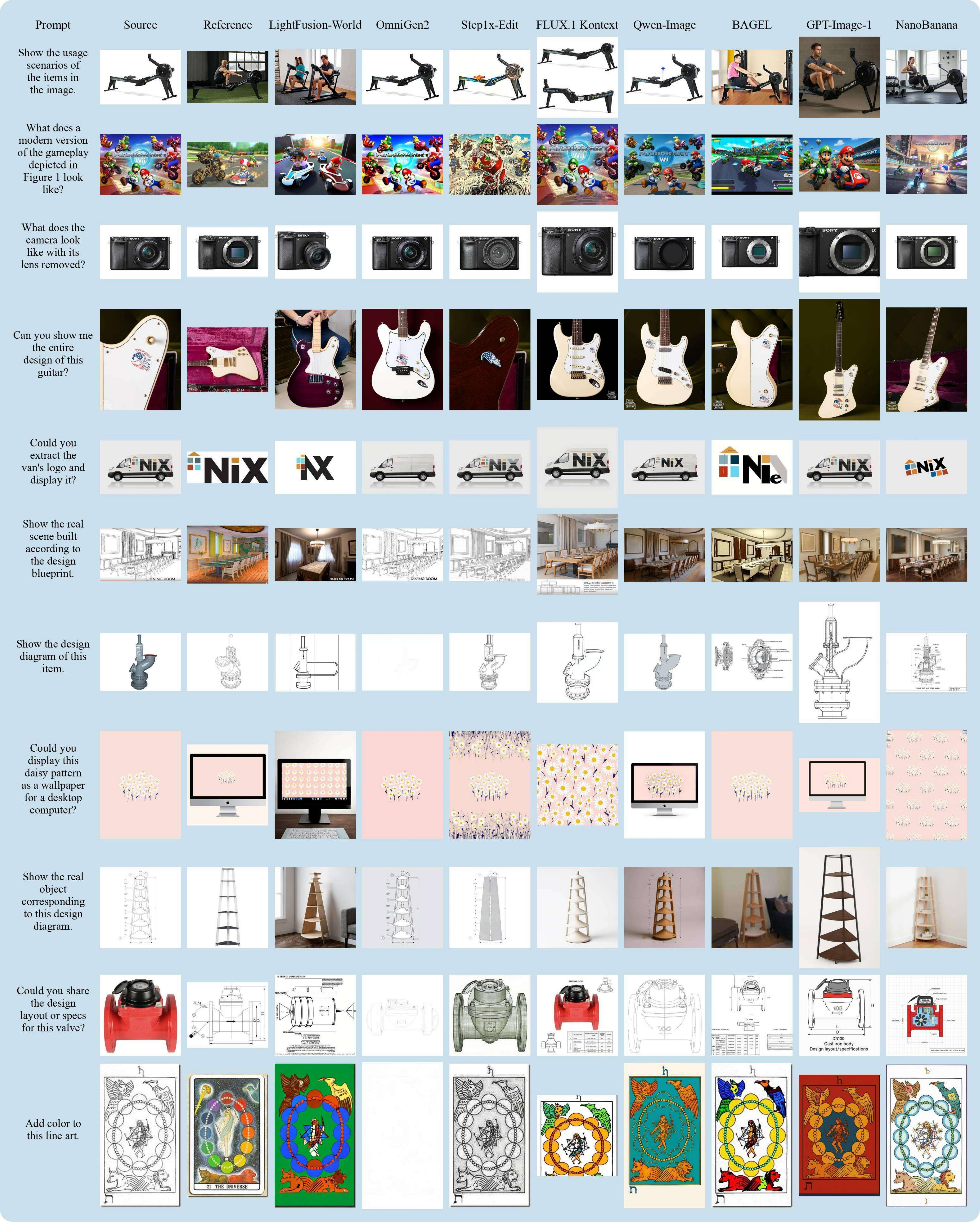}
    \caption{Comprehensive visualization of model performance on IntelligentBench (Subset Design, part 5/9).}\label{fig:sub:design:5}
\end{figure*}

\begin{figure*}[ht]
    \centering
    \includegraphics[width=\linewidth]{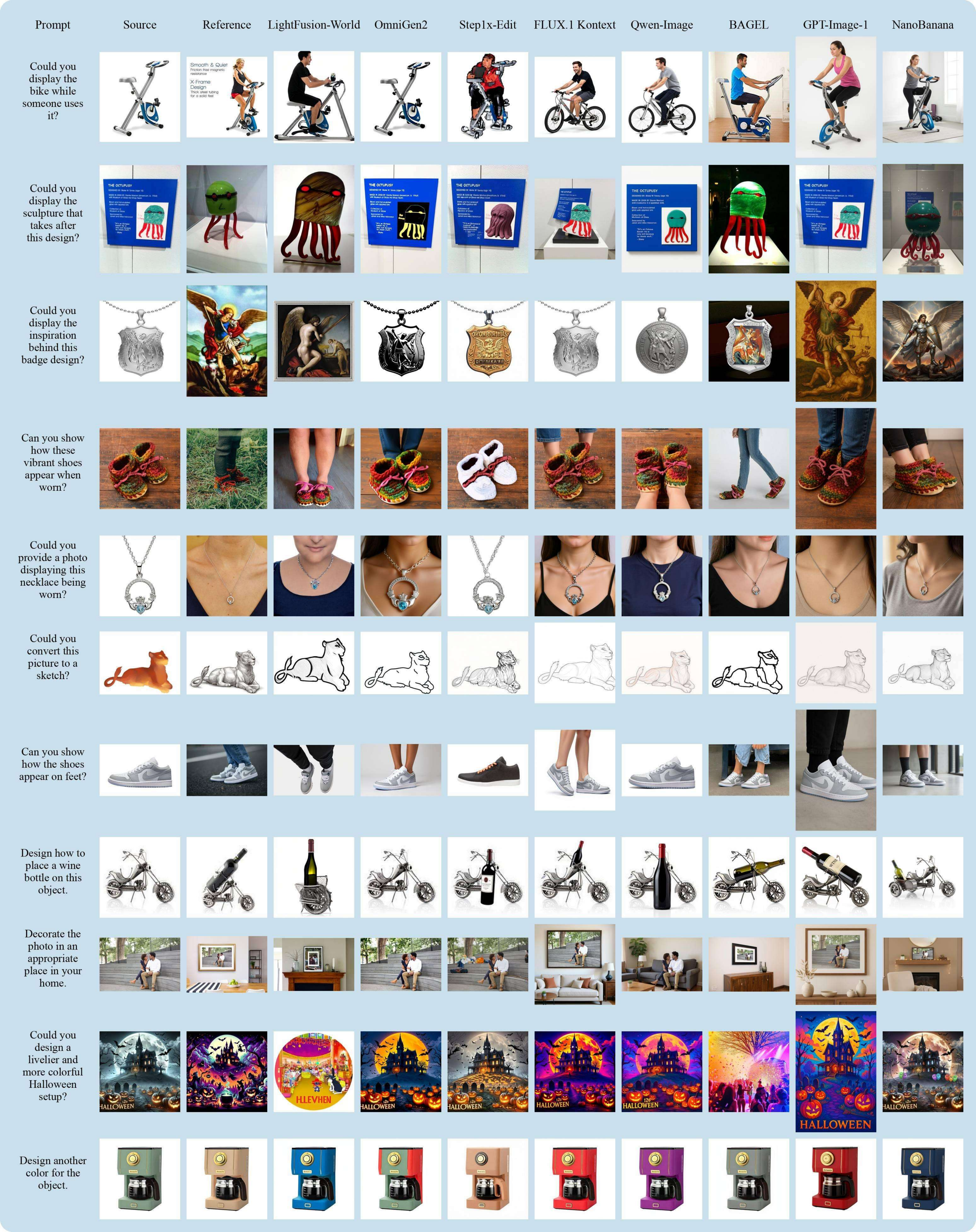}
    \caption{Comprehensive visualization of model performance on IntelligentBench (Subset Design, part 6/9).}\label{fig:sub:design:6}
\end{figure*}

\begin{figure*}[ht]
    \centering
    \includegraphics[width=\linewidth]{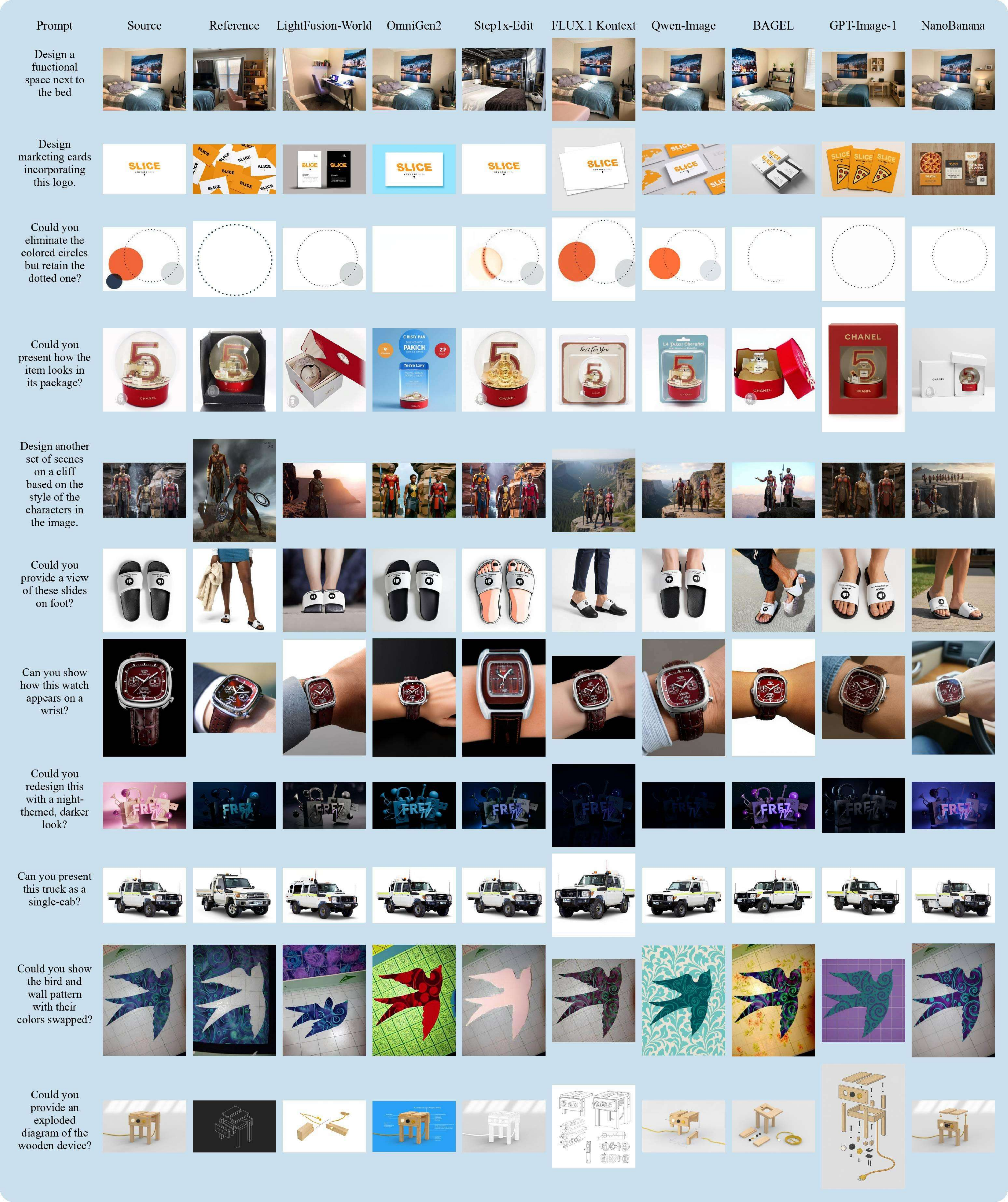}
    \caption{Comprehensive visualization of model performance on IntelligentBench (Subset Design, part 7/9).}\label{fig:sub:design:7}
\end{figure*}

\begin{figure*}[ht]
    \centering
    \includegraphics[width=\linewidth]{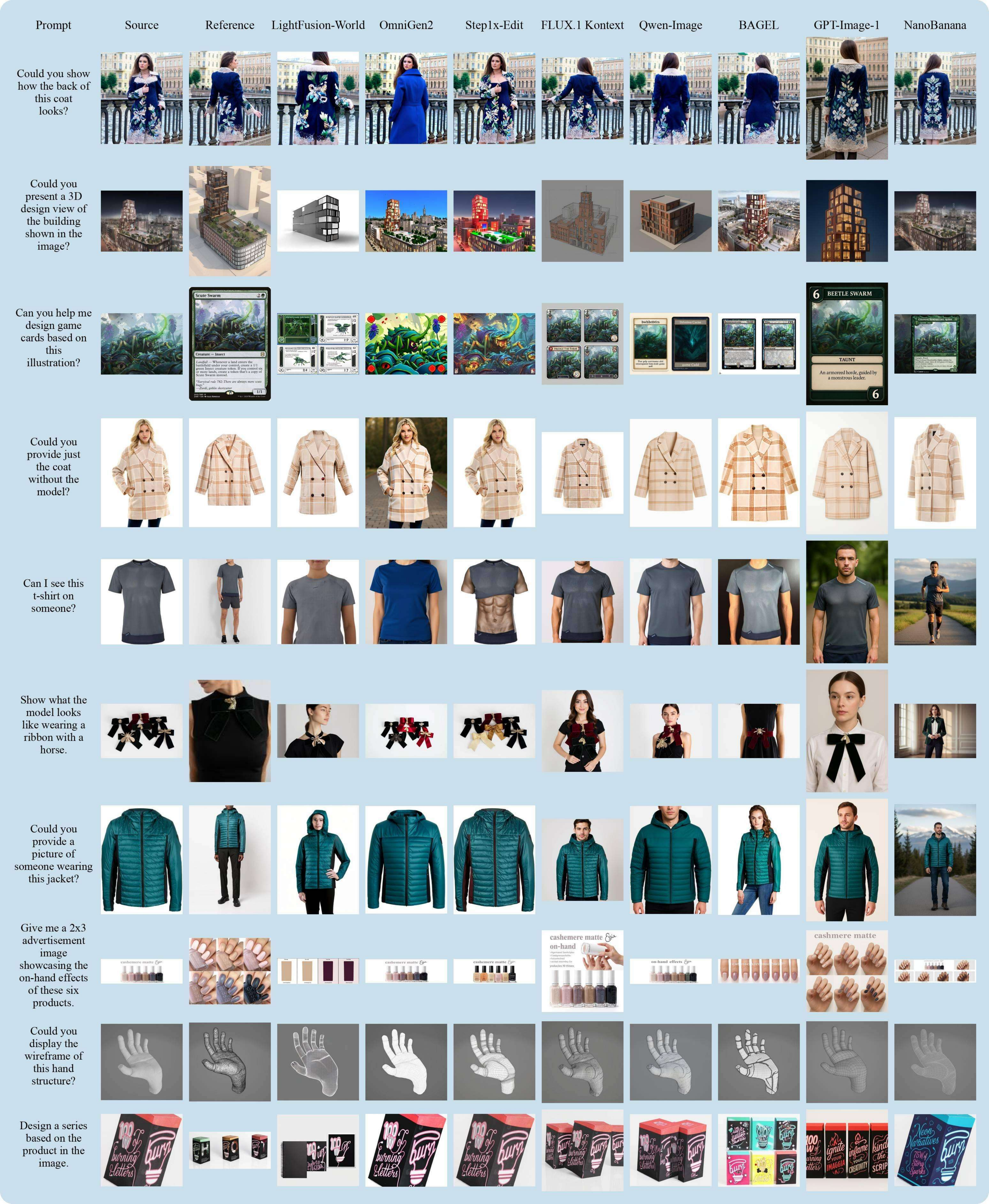}
    \caption{Comprehensive visualization of model performance on IntelligentBench (Subset Design, part 8/9).}\label{fig:sub:design:8}
\end{figure*}

\begin{figure*}[ht]
    \centering
    \includegraphics[width=\linewidth]{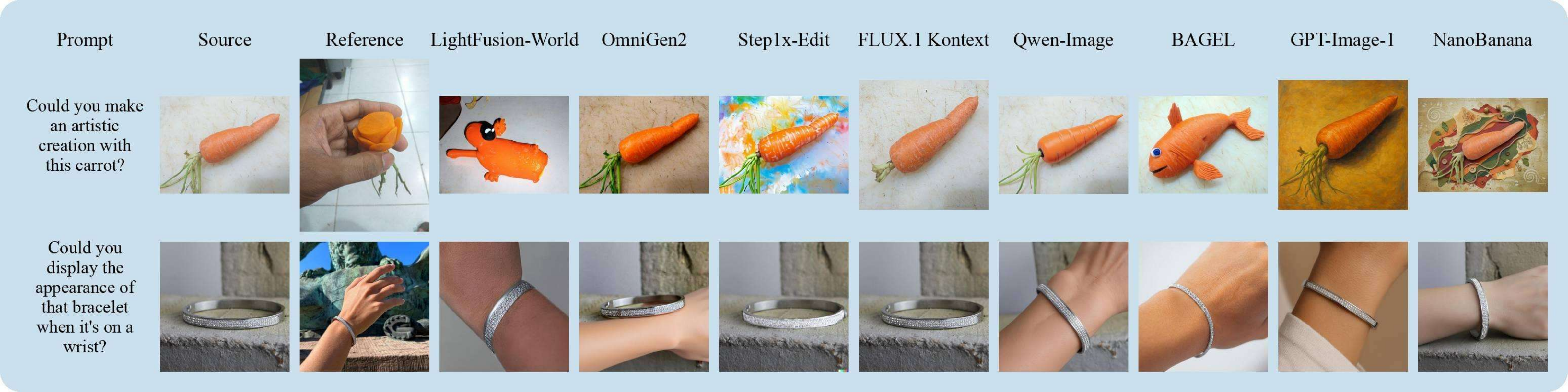}
    \caption{Comprehensive visualization of model performance on IntelligentBench (Subset Design, part 9/9).}\label{fig:sub:design:9}
\end{figure*}

\begin{figure*}[ht]
    \centering
    \includegraphics[width=\linewidth]{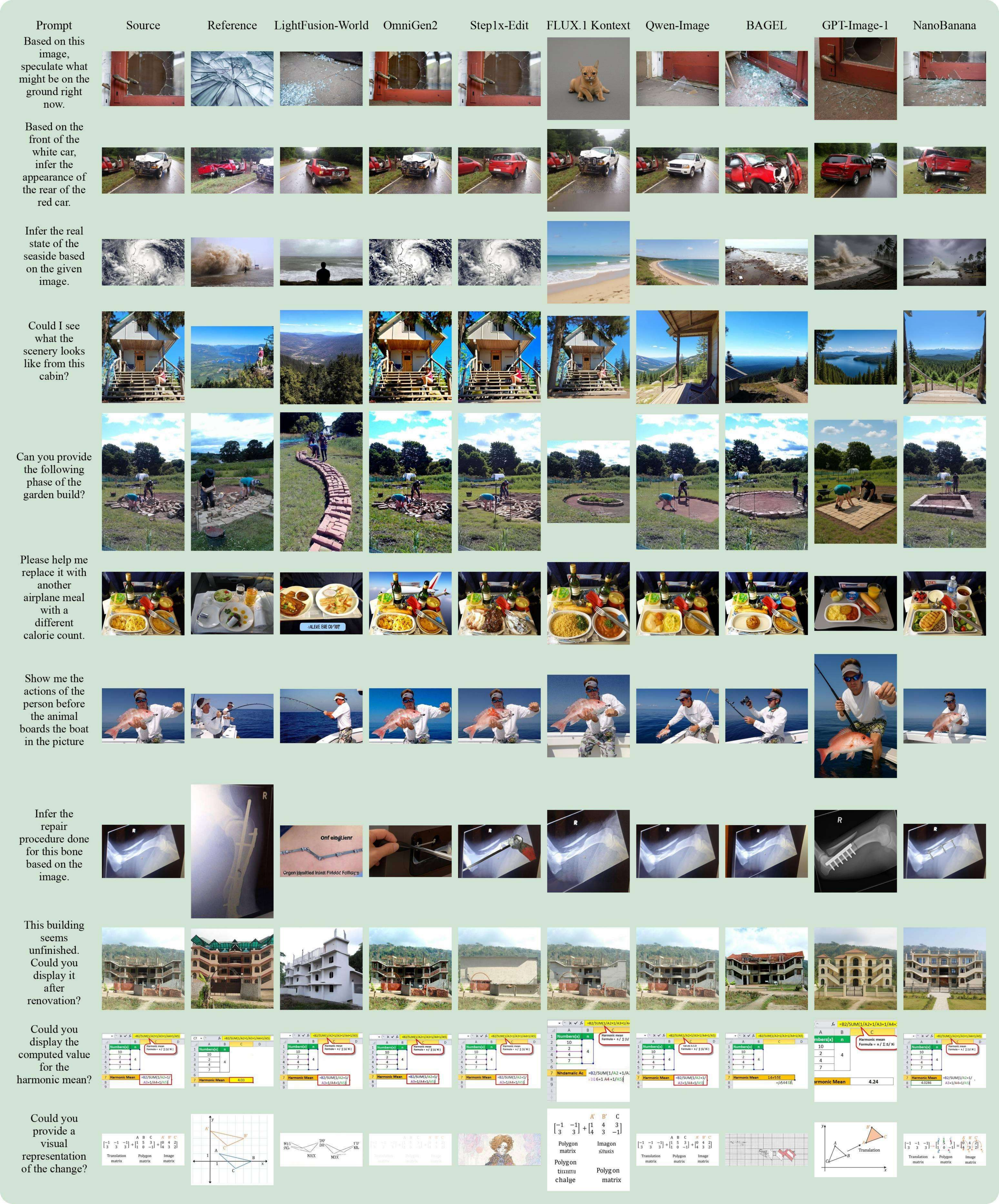}
    \caption{Comprehensive visualization of model performance on IntelligentBench (Subset Reasoning, part 1/9).}\label{fig:sub:reasoning:1}
\end{figure*}

\begin{figure*}[ht]
    \centering
    \includegraphics[width=\linewidth]{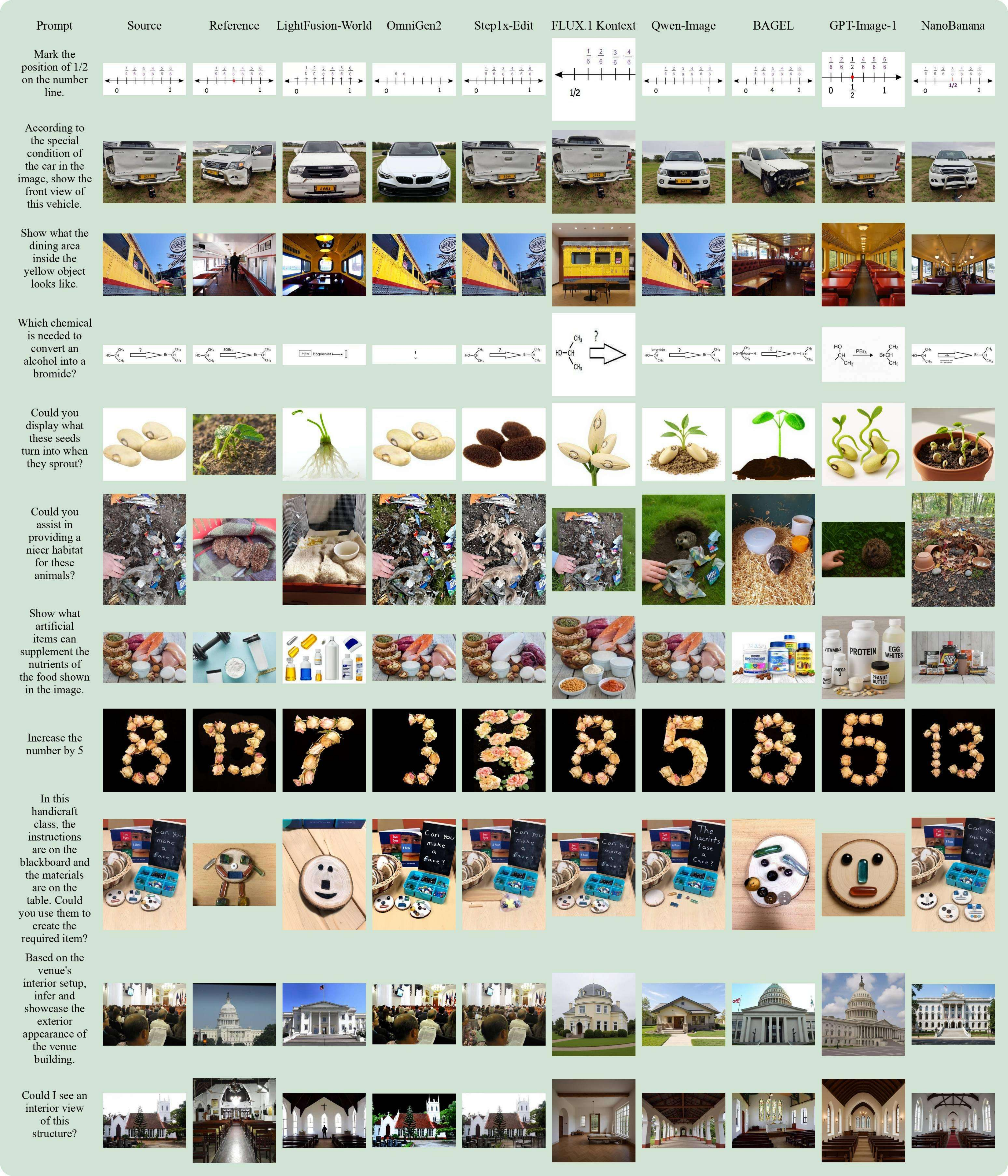}
    \caption{Comprehensive visualization of model performance on IntelligentBench (Subset Reasoning, part 2/9).}\label{fig:sub:reasoning:2}
\end{figure*}

\begin{figure*}[ht]
    \centering
    \includegraphics[width=\linewidth]{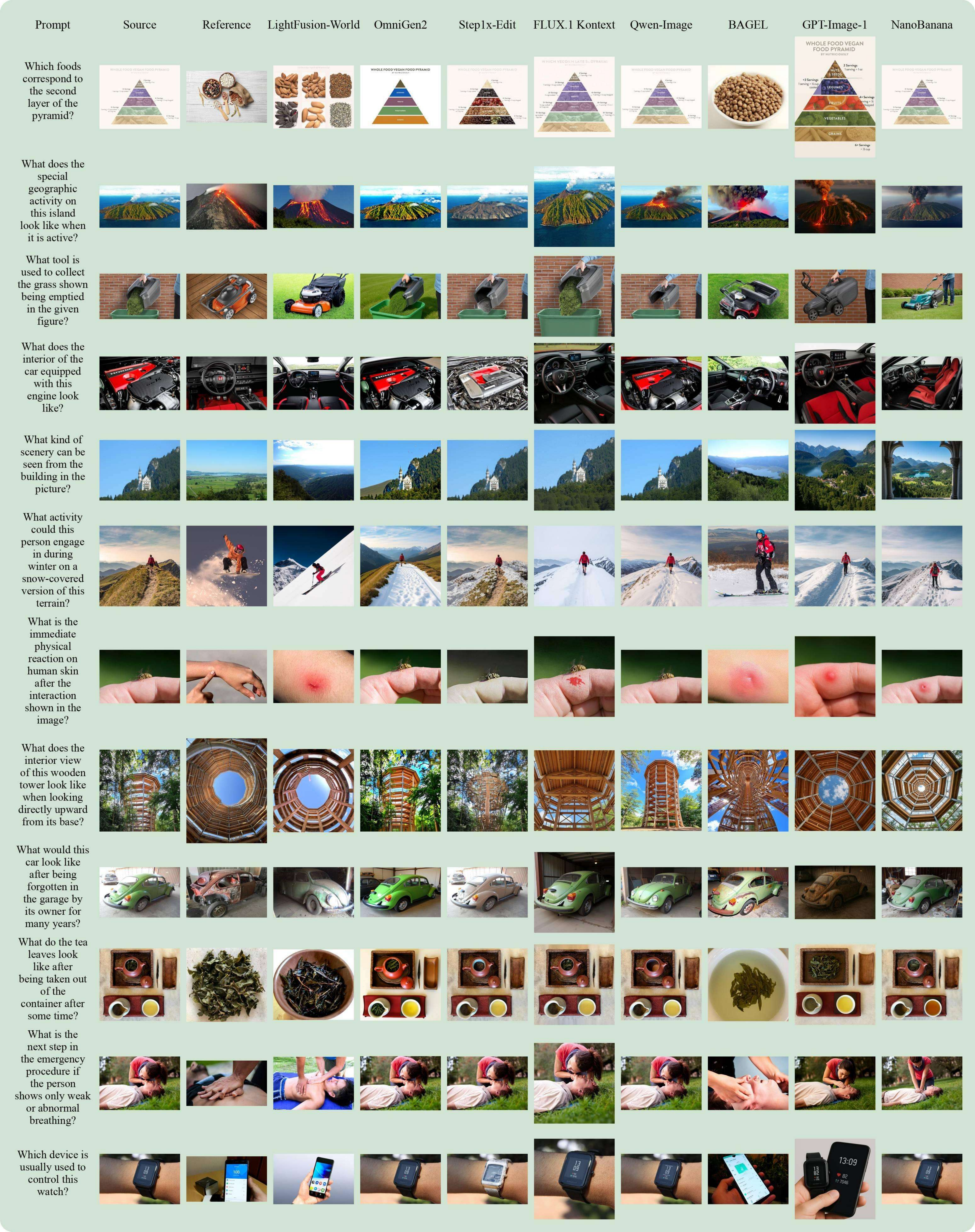}
    \caption{Comprehensive visualization of model performance on IntelligentBench (Subset Reasoning, part 3/9).}\label{fig:sub:reasoning:3}
\end{figure*}

\begin{figure*}[ht]
    \centering
    \includegraphics[width=\linewidth]{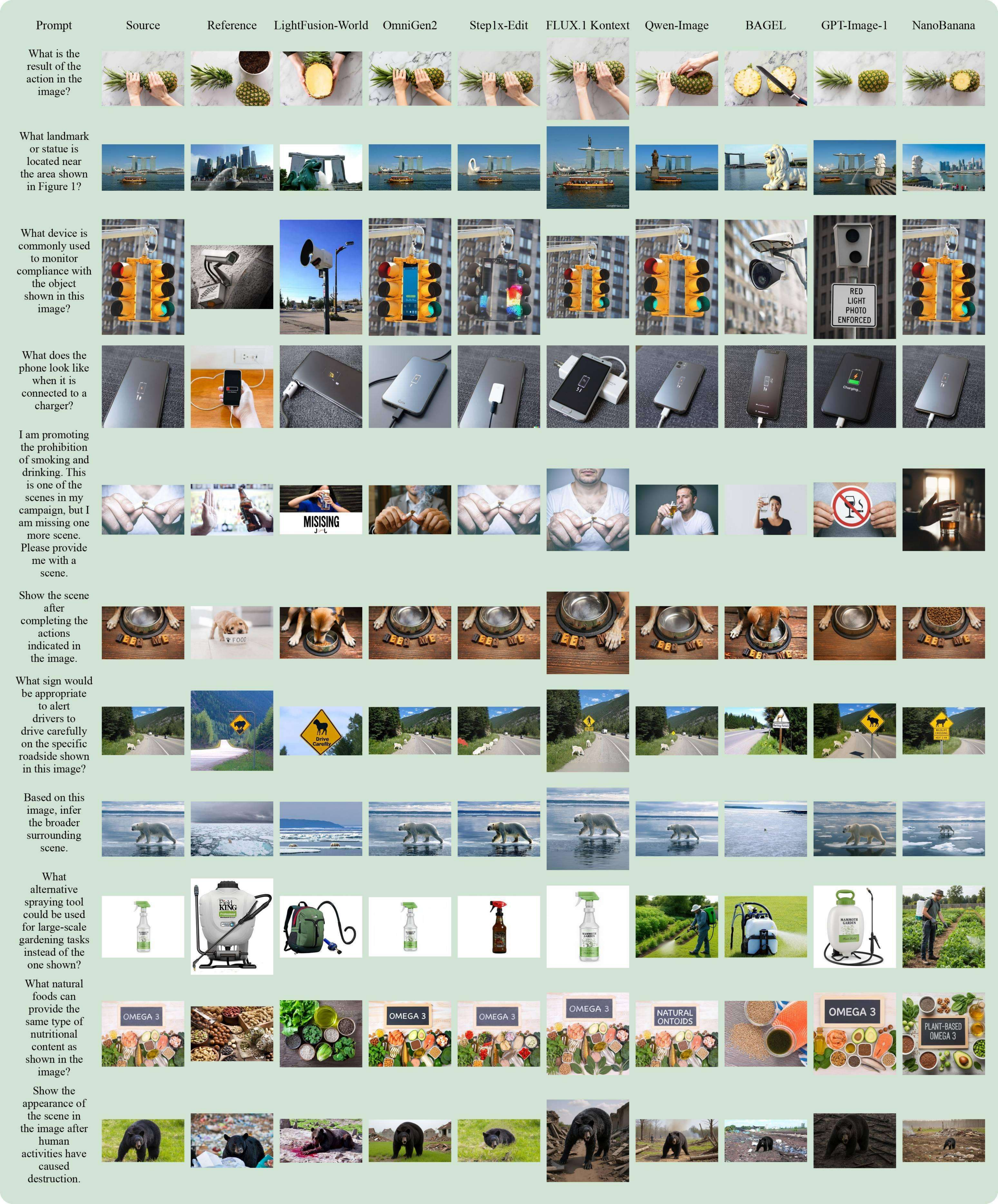}
    \caption{Comprehensive visualization of model performance on IntelligentBench (Subset Reasoning, part 4/9).}\label{fig:sub:reasoning:4}
\end{figure*}

\begin{figure*}[ht]
    \centering
    \includegraphics[width=\linewidth]{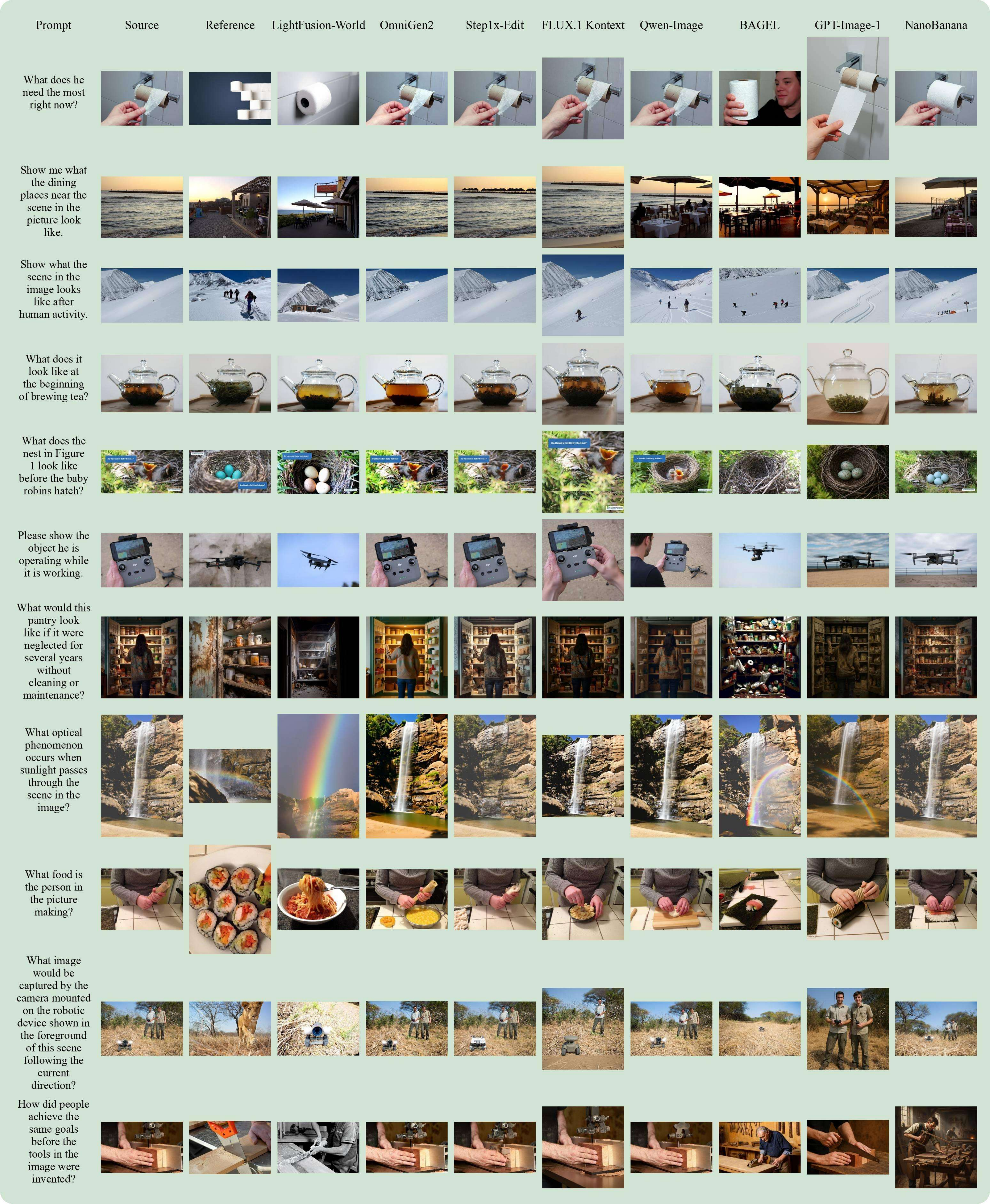}
    \caption{Comprehensive visualization of model performance on IntelligentBench (Subset Reasoning, part 5/9).}\label{fig:sub:reasoning:5}
\end{figure*}

\begin{figure*}[ht]
    \centering
    \includegraphics[width=\linewidth]{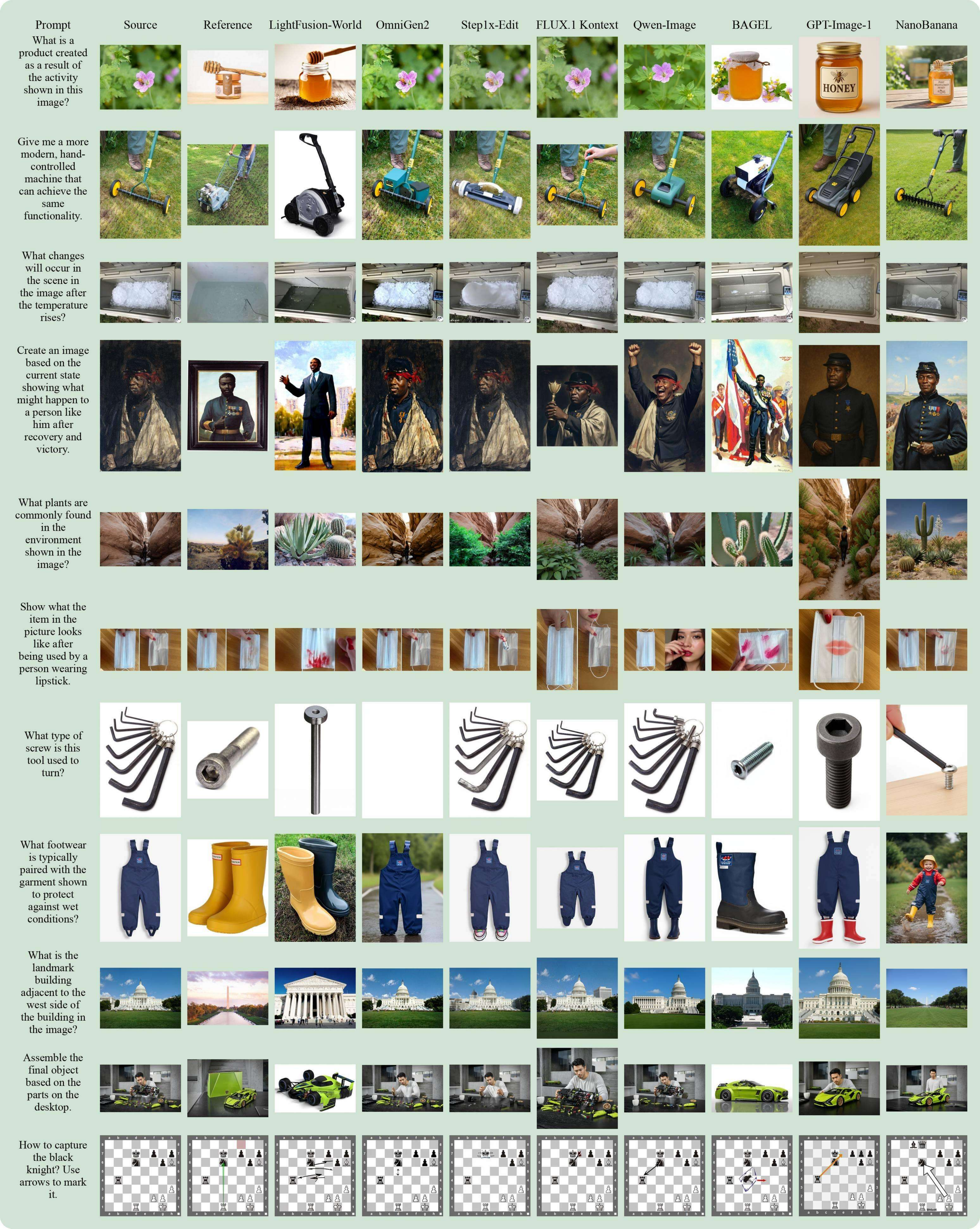}
    \caption{Comprehensive visualization of model performance on IntelligentBench (Subset Reasoning, part 6/9).}\label{fig:sub:reasoning:6}
\end{figure*}

\begin{figure*}[ht]
    \centering
    \includegraphics[width=\linewidth]{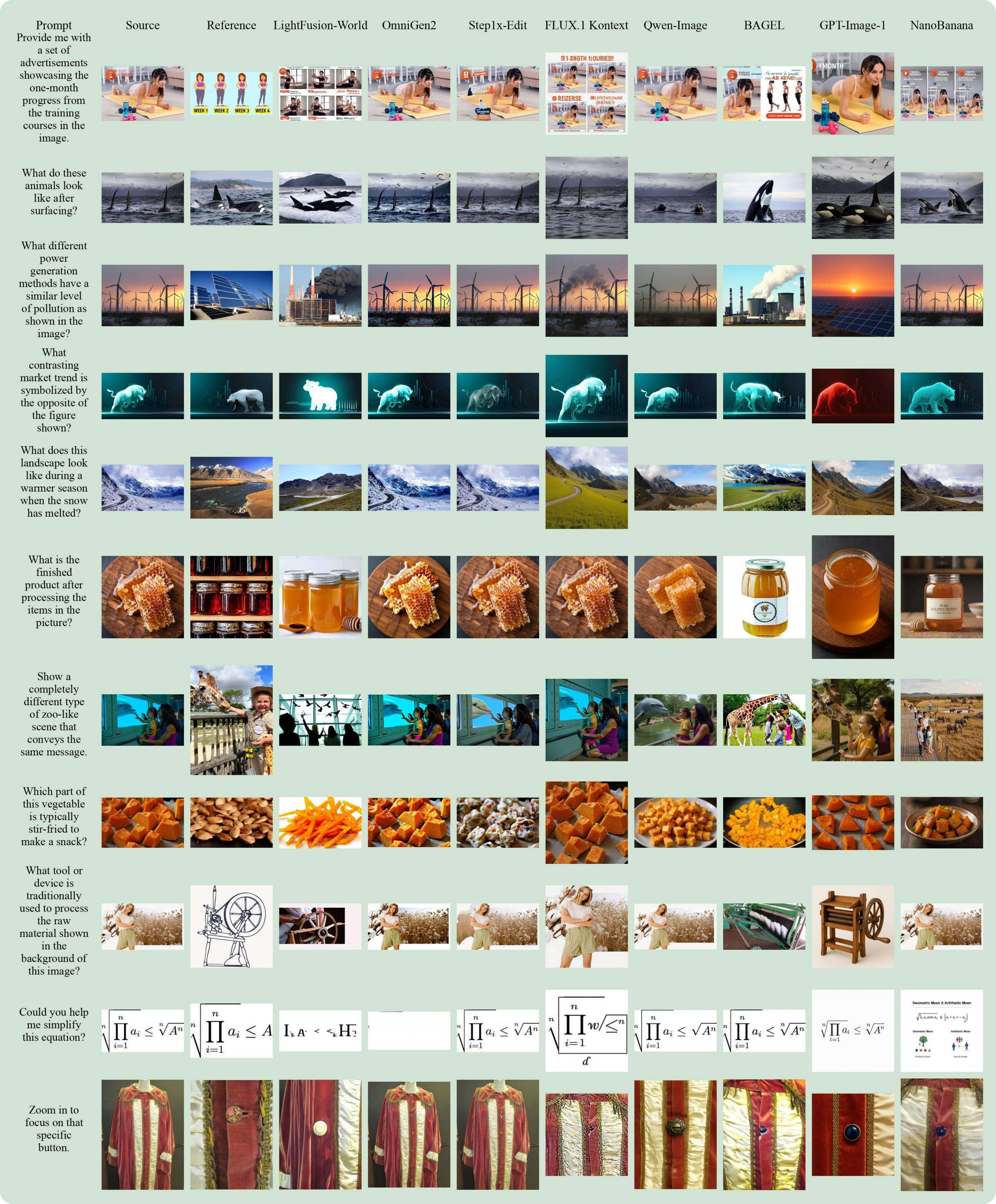}
    \caption{Comprehensive visualization of model performance on IntelligentBench (Subset Reasoning, part 7/9).}\label{fig:sub:reasoning:7}
\end{figure*}

\begin{figure*}[ht]
    \centering
    \includegraphics[width=\linewidth]{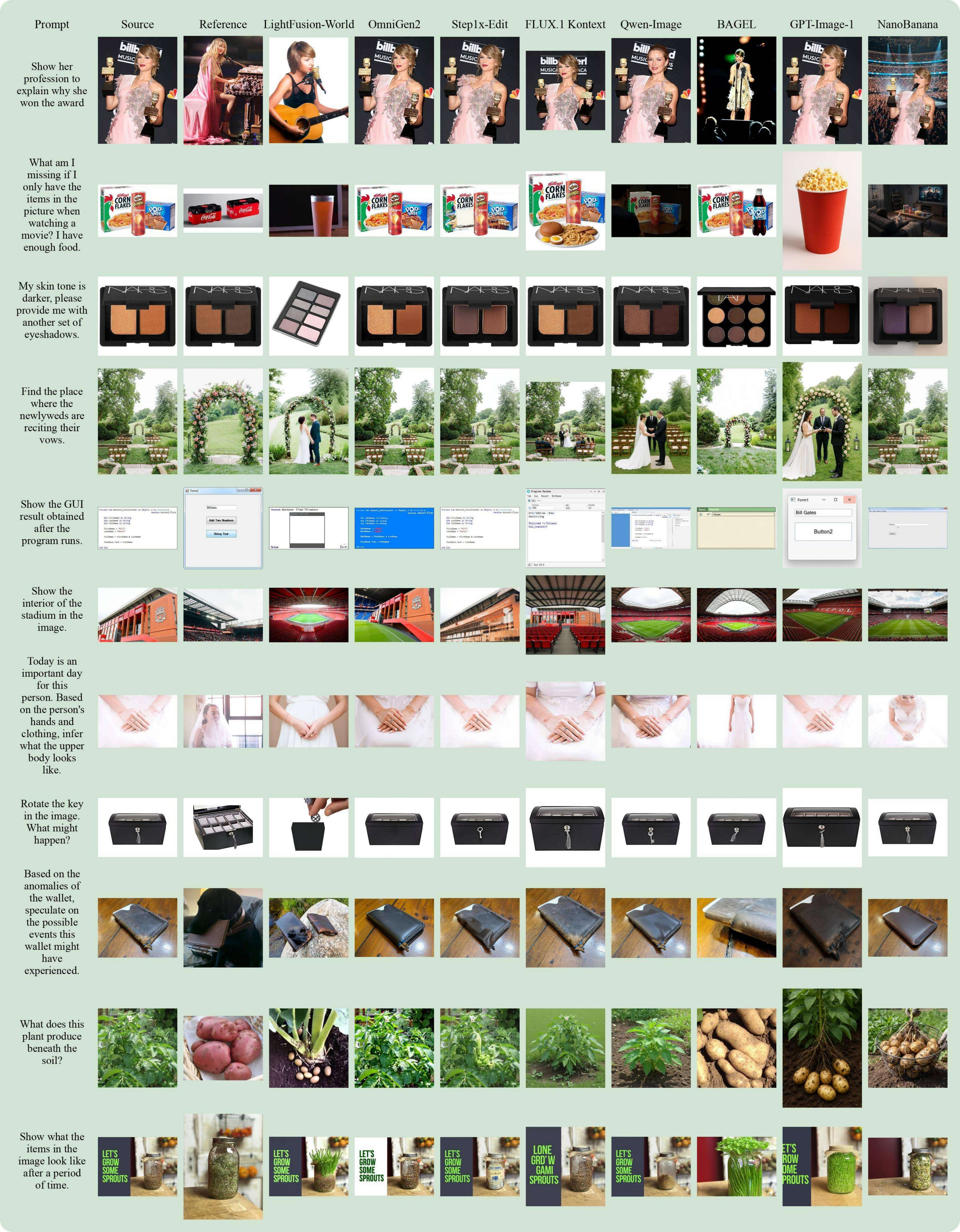}
    \caption{Comprehensive visualization of model performance on IntelligentBench (Subset Reasoning, part 8/9).}\label{fig:sub:reasoning:8}
\end{figure*}

\begin{figure*}[ht]
    \centering
    \includegraphics[width=\linewidth]{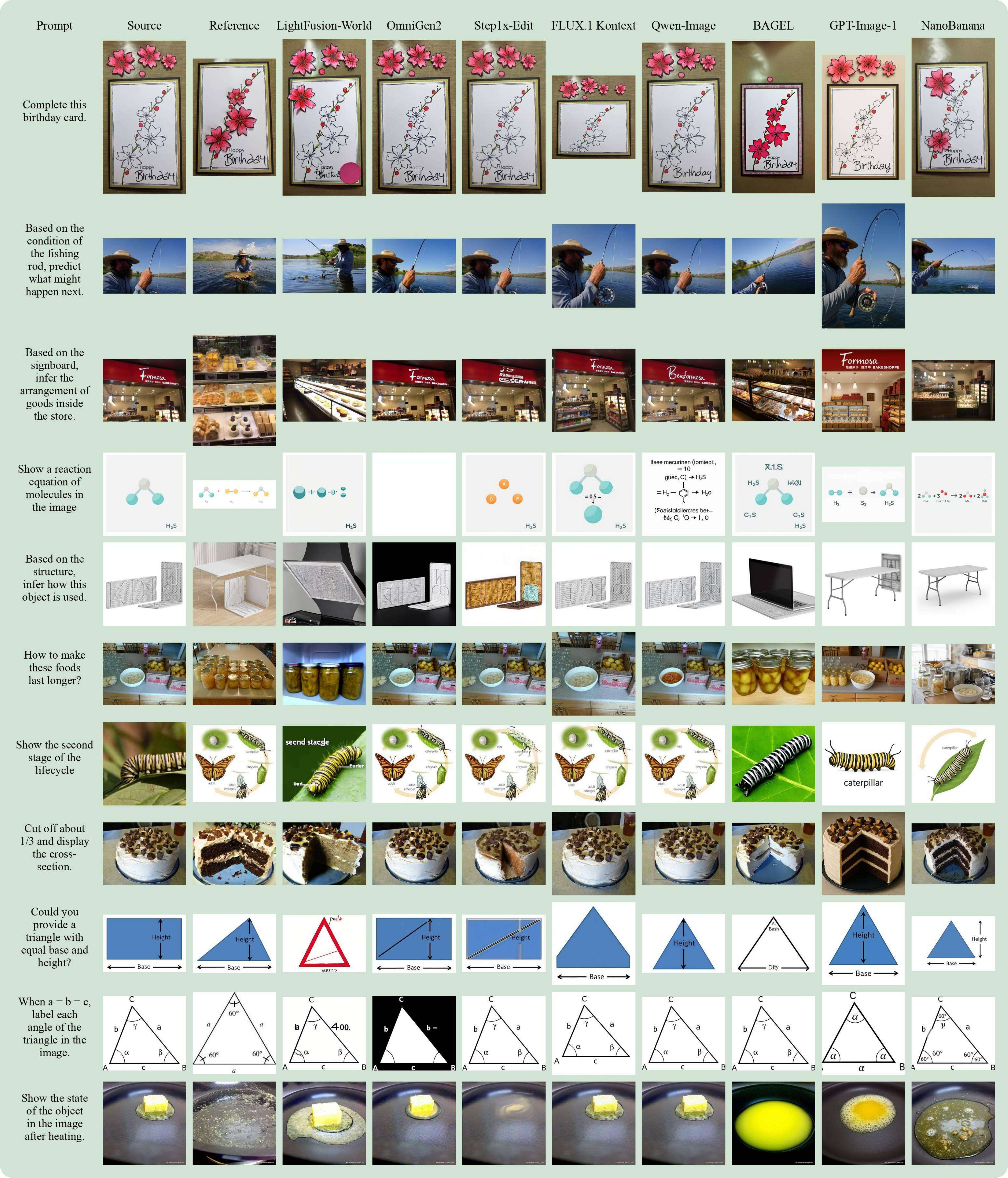}
    \caption{Comprehensive visualization of model performance on IntelligentBench (Subset Reasoning, part 9/9).}\label{fig:sub:reasoning:9}
\end{figure*}

\begin{figure*}[ht]
    \centering
    \includegraphics[width=\linewidth]{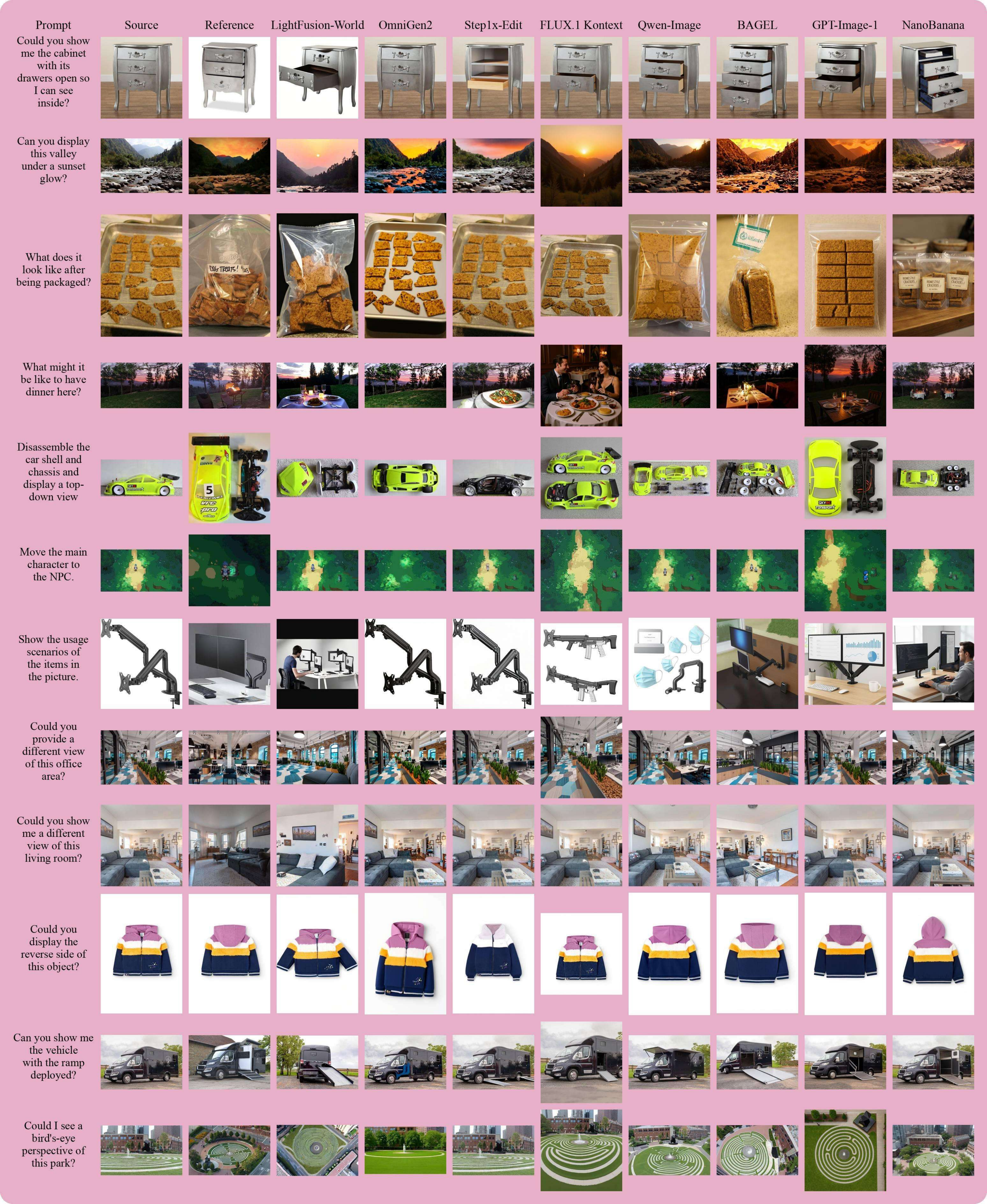}
    \caption{Comprehensive visualization of model performance on IntelligentBench (Subset World knowledge, part 1/15).}\label{fig:sub:knowledge:1}
\end{figure*}

\begin{figure*}[ht]
    \centering
    \includegraphics[width=\linewidth]{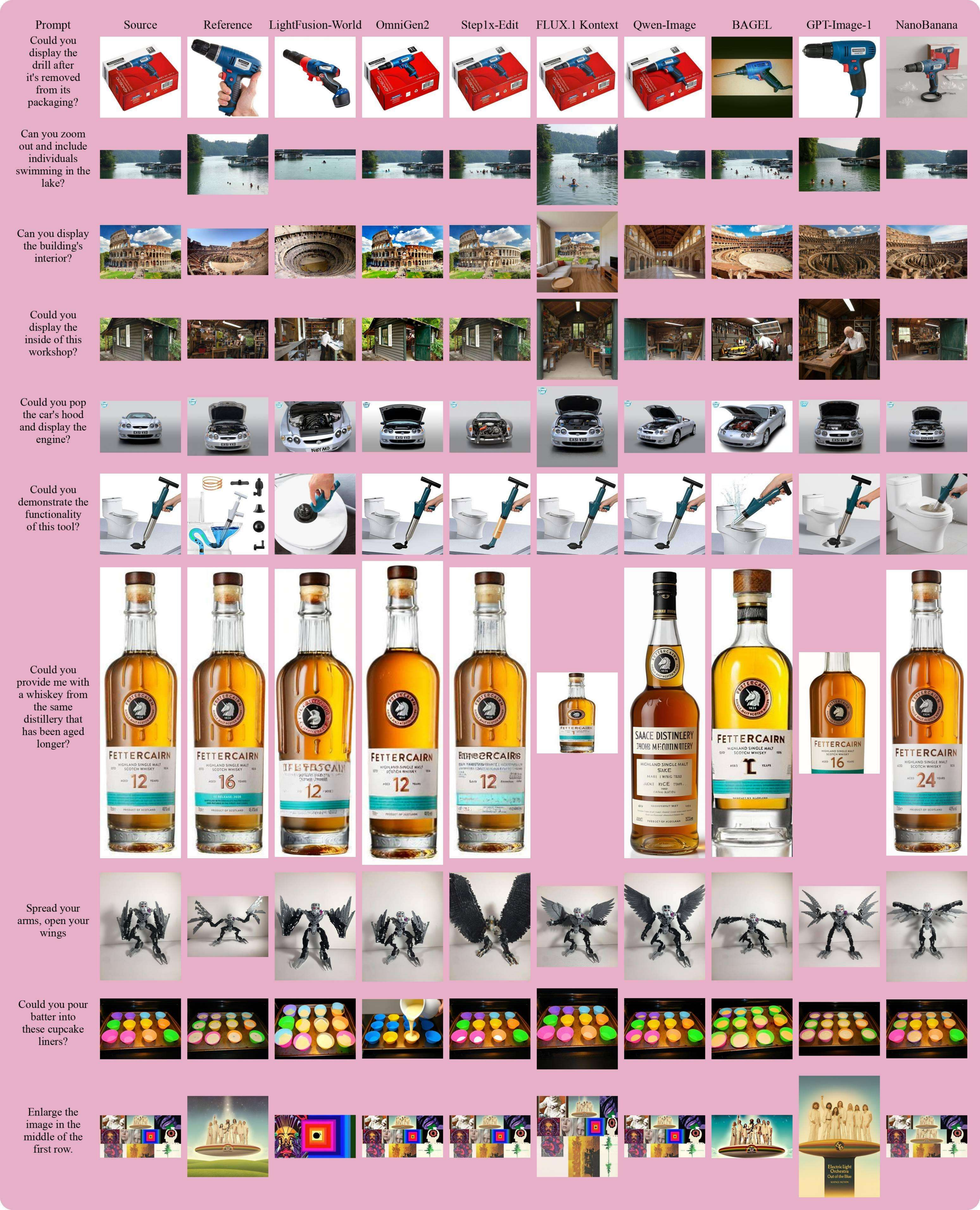}
    \caption{Comprehensive visualization of model performance on IntelligentBench (Subset World knowledge, part 2/15).}\label{fig:sub:knowledge:2}
\end{figure*}

\begin{figure*}[ht]
    \centering
    \includegraphics[width=\linewidth]{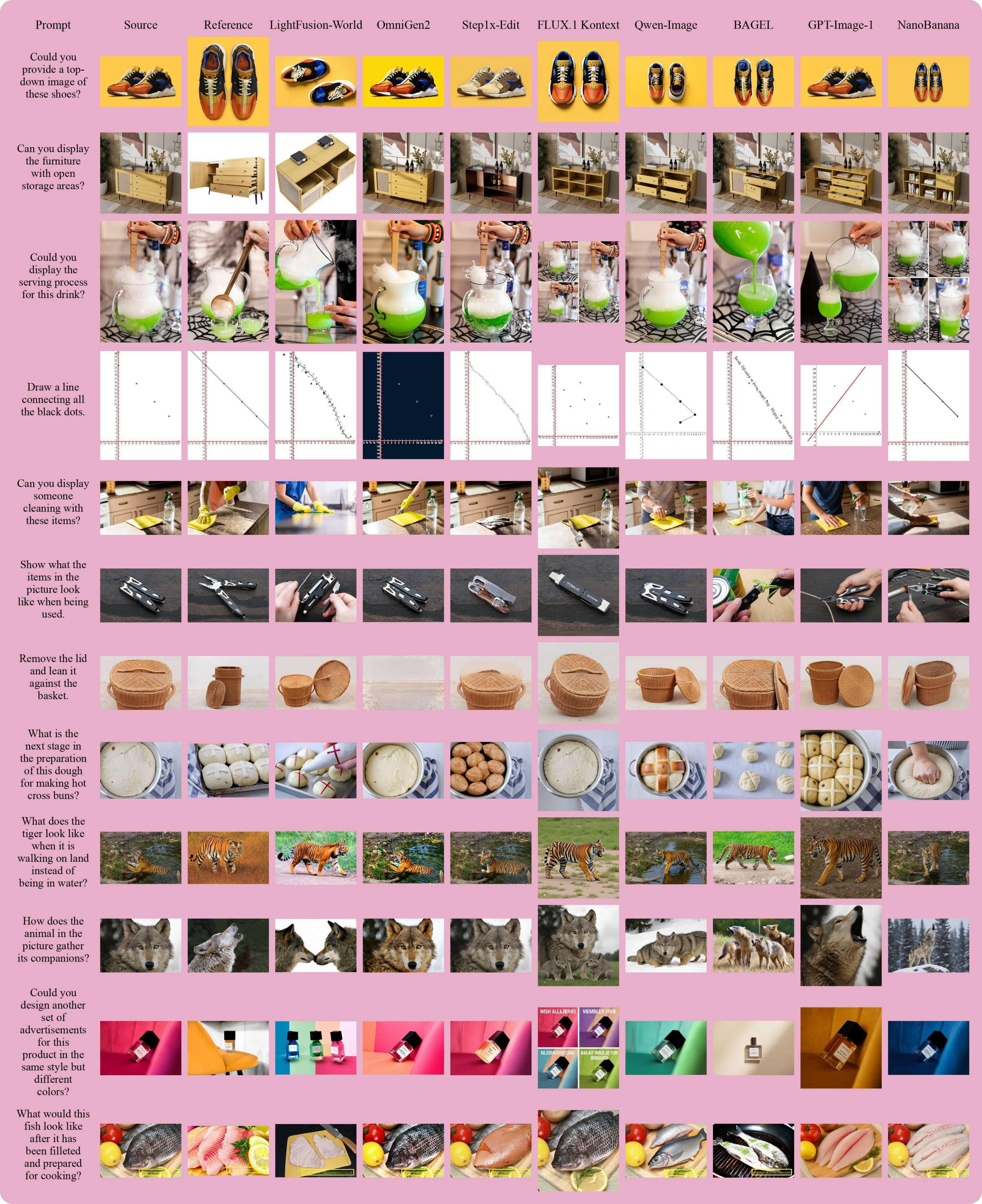}
    \caption{Comprehensive visualization of model performance on IntelligentBench (Subset World knowledge, part 3/15).}\label{fig:sub:knowledge:3}
\end{figure*}

\begin{figure*}[ht]
    \centering
    \includegraphics[width=\linewidth]{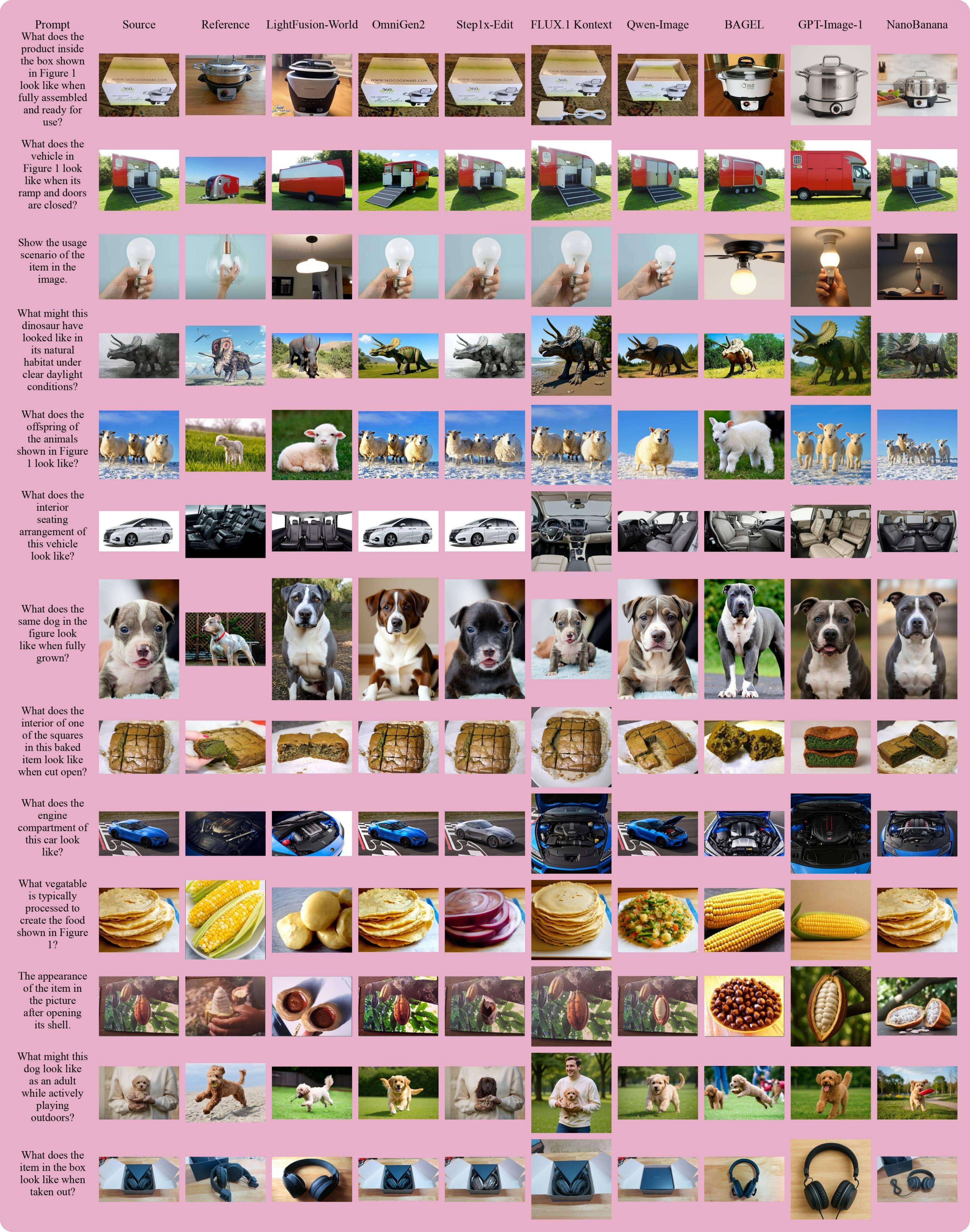}
    \caption{Comprehensive visualization of model performance on IntelligentBench (Subset World knowledge, part 4/15).}\label{fig:sub:knowledge:4}
\end{figure*}

\begin{figure*}[ht]
    \centering
    \includegraphics[width=\linewidth]{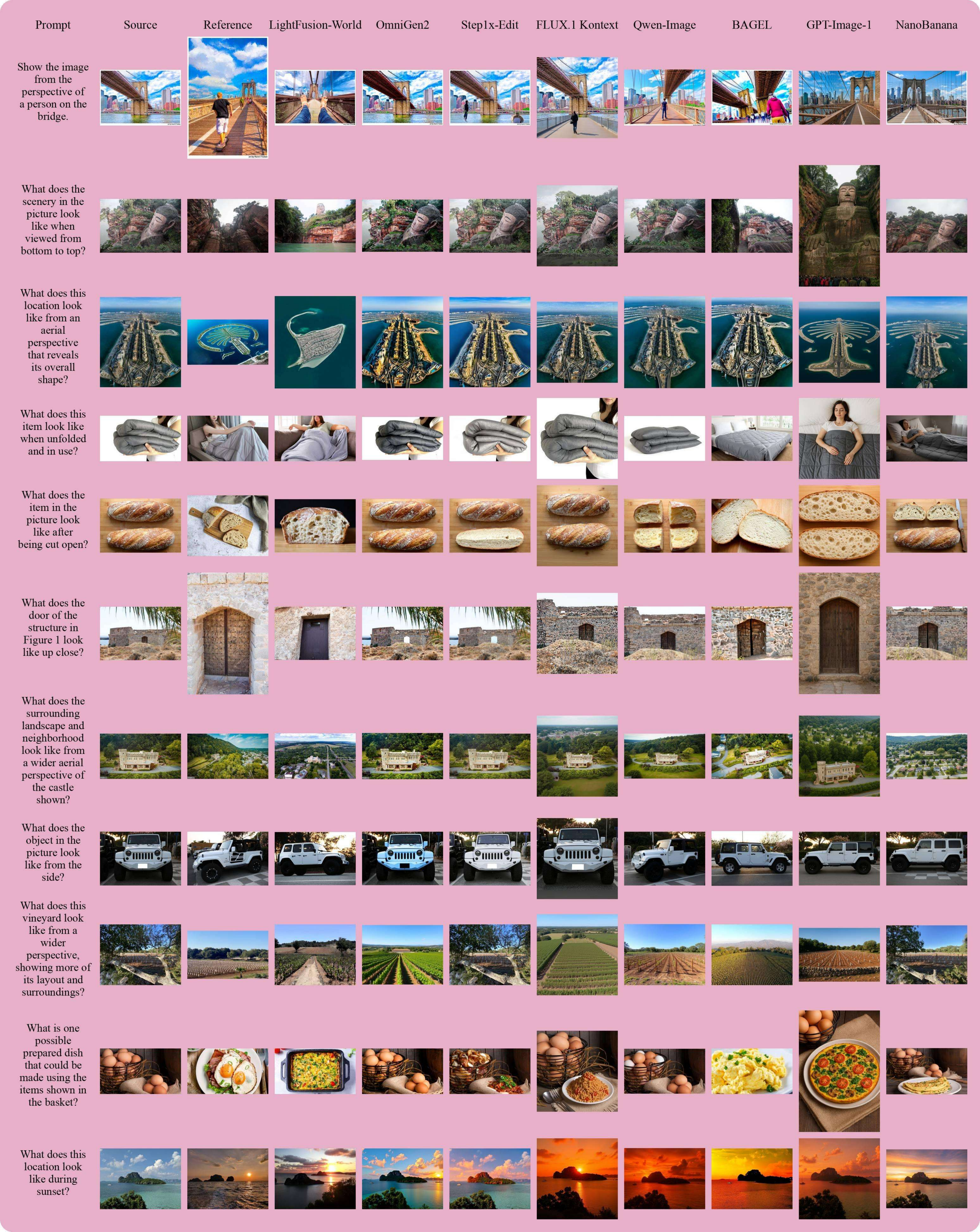}
    \caption{Comprehensive visualization of model performance on IntelligentBench (Subset World knowledge, part 5/15).}\label{fig:sub:knowledge:5}
\end{figure*}

\begin{figure*}[ht]
    \centering
    \includegraphics[width=\linewidth]{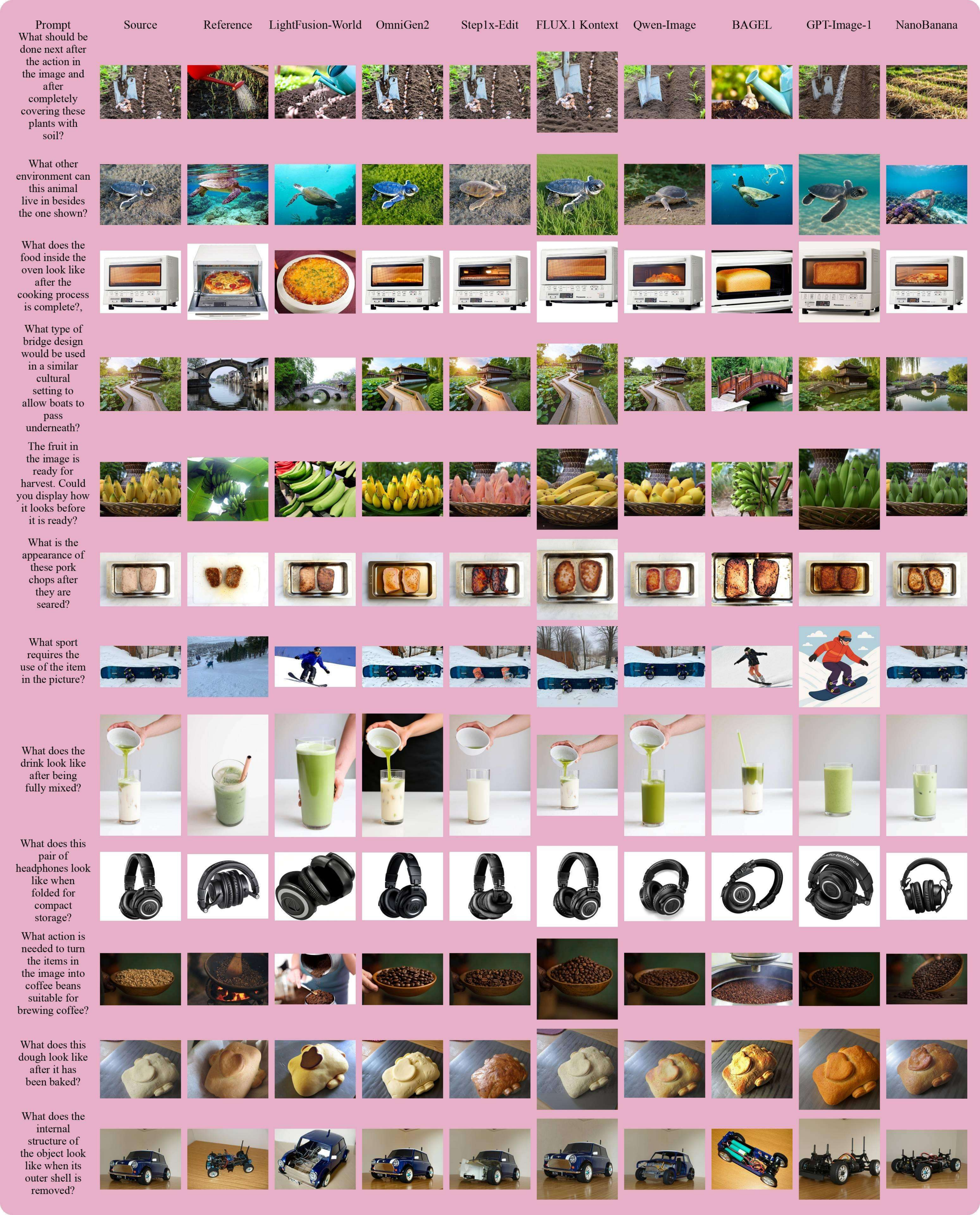}
    \caption{Comprehensive visualization of model performance on IntelligentBench (Subset World knowledge, part 6/15).}\label{fig:sub:knowledge:6}
\end{figure*}

\begin{figure*}[ht]
    \centering
    \includegraphics[width=\linewidth]{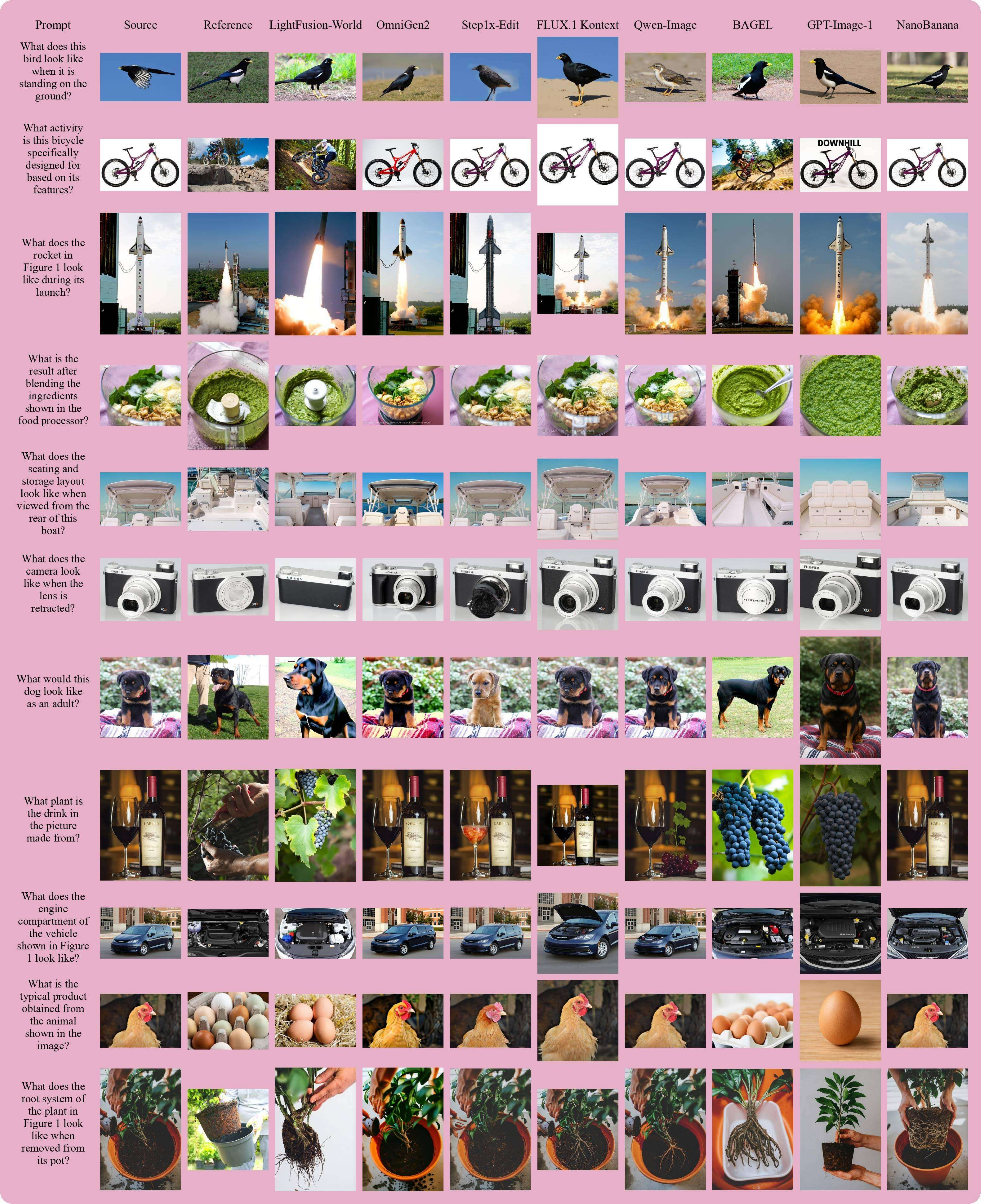}
    \caption{Comprehensive visualization of model performance on IntelligentBench (Subset World knowledge, part 7/15).}\label{fig:sub:knowledge:7}
\end{figure*}

\begin{figure*}[ht]
    \centering
    \includegraphics[width=\linewidth]{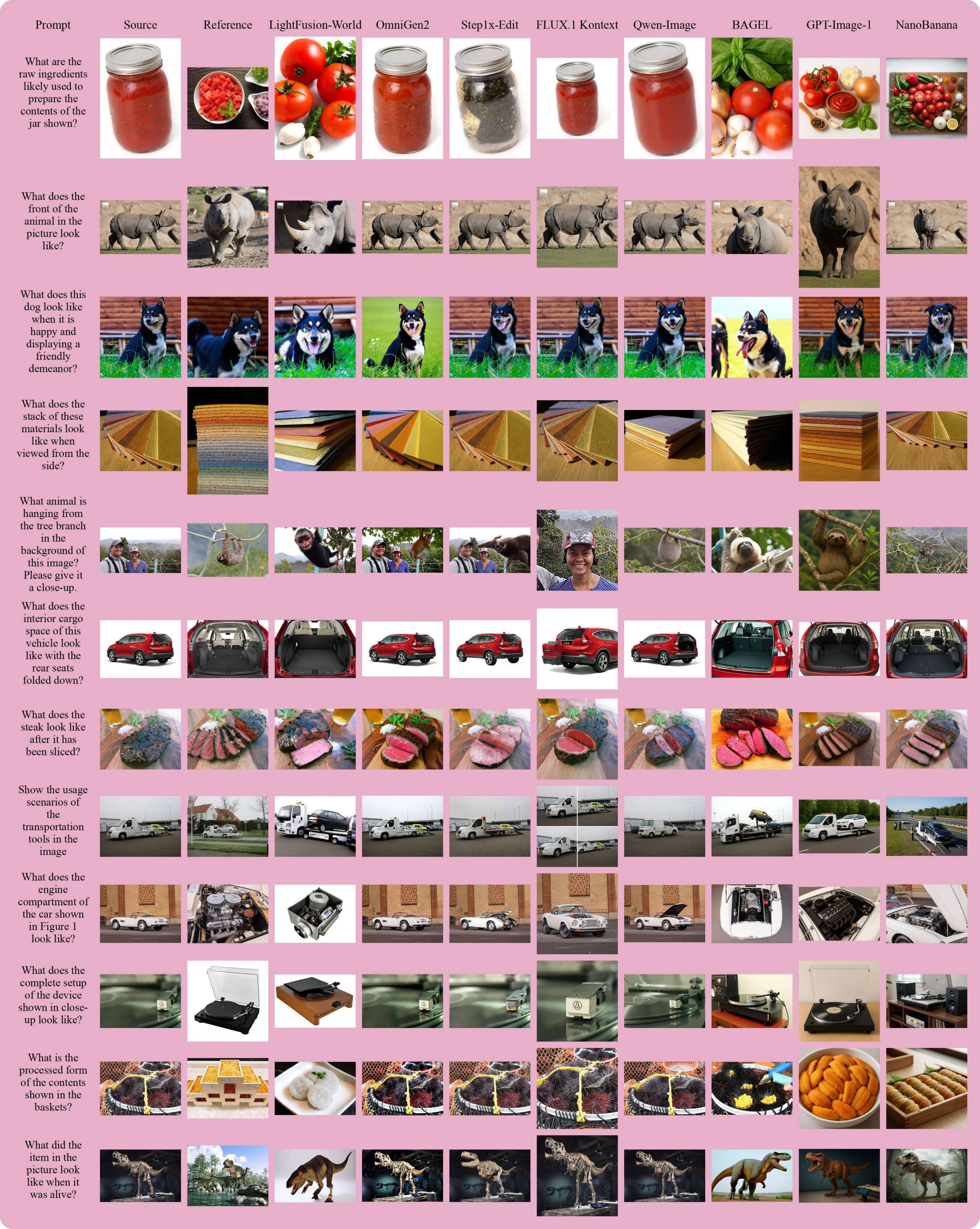}
    \caption{Comprehensive visualization of model performance on IntelligentBench (Subset World knowledge, part 8/15).}\label{fig:sub:knowledge:8}
\end{figure*}

\begin{figure*}[ht]
    \centering
    \includegraphics[width=\linewidth]{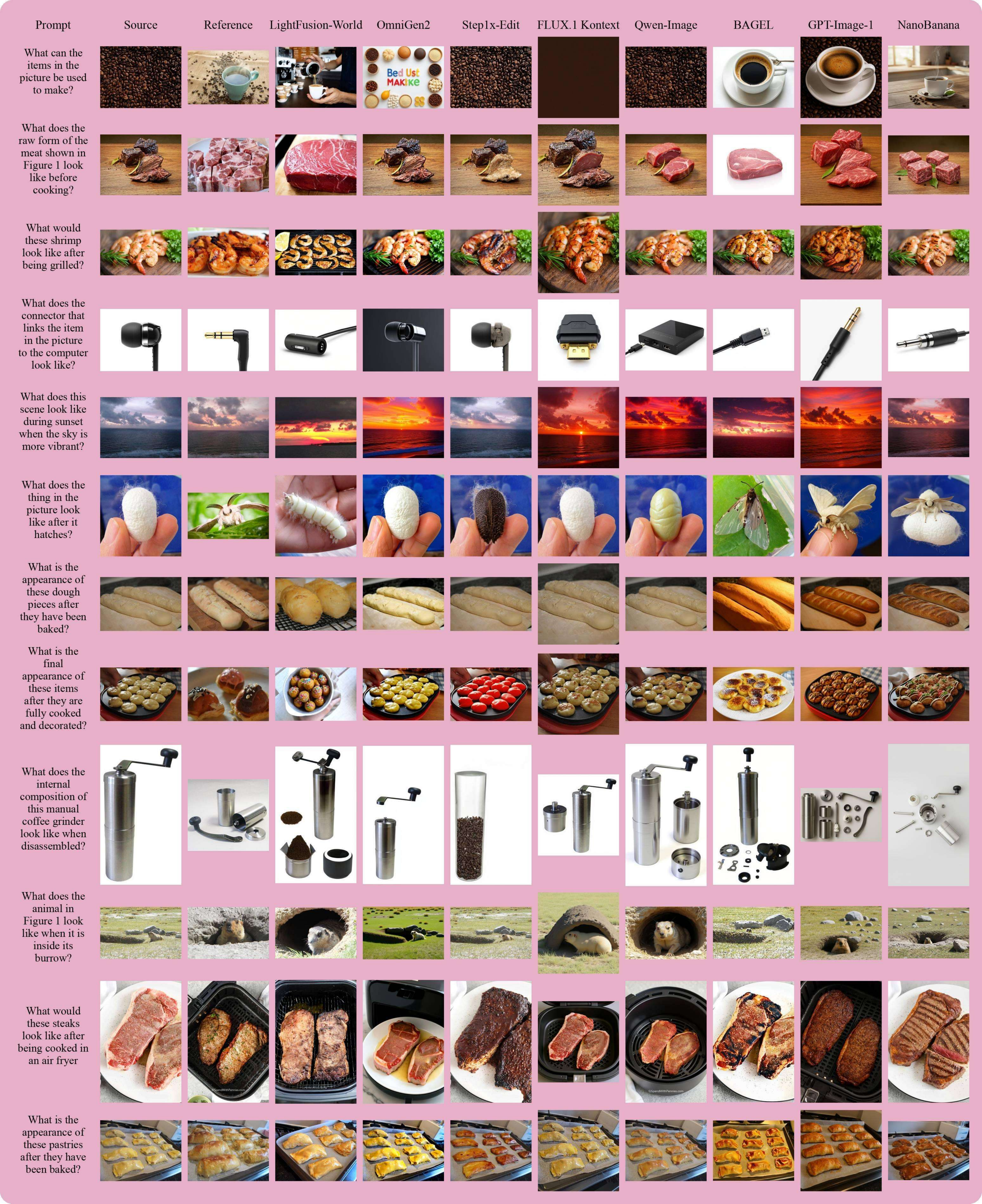}
    \caption{Comprehensive visualization of model performance on IntelligentBench (Subset World knowledge, part 9/15).}\label{fig:sub:knowledge:9}
\end{figure*}

\begin{figure*}[ht]
    \centering
    \includegraphics[width=\linewidth]{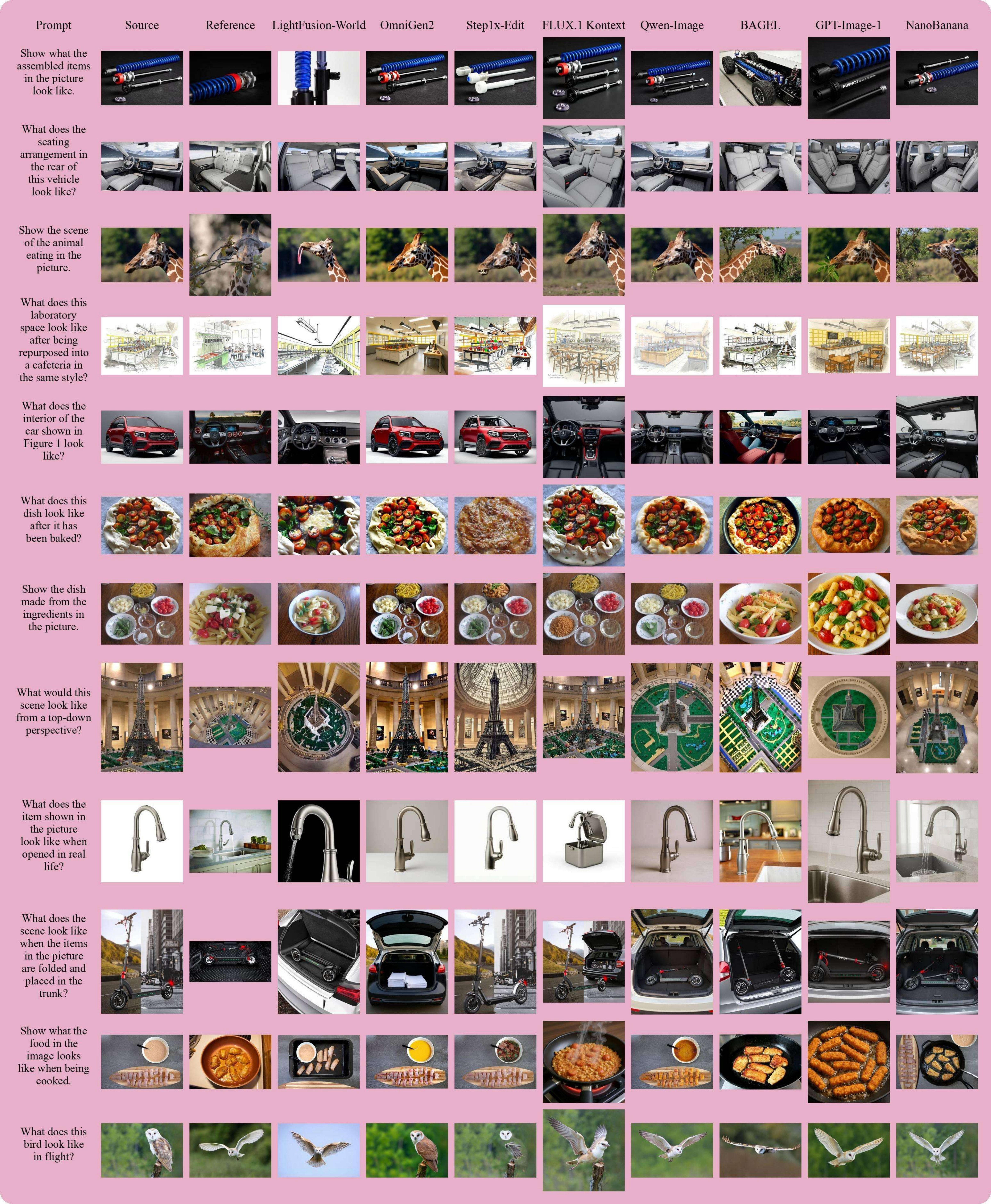}
    \caption{Comprehensive visualization of model performance on IntelligentBench (Subset World knowledge, part 10/15).}\label{fig:sub:knowledge:10}
\end{figure*}

\begin{figure*}[ht]
    \centering
    \includegraphics[width=\linewidth]{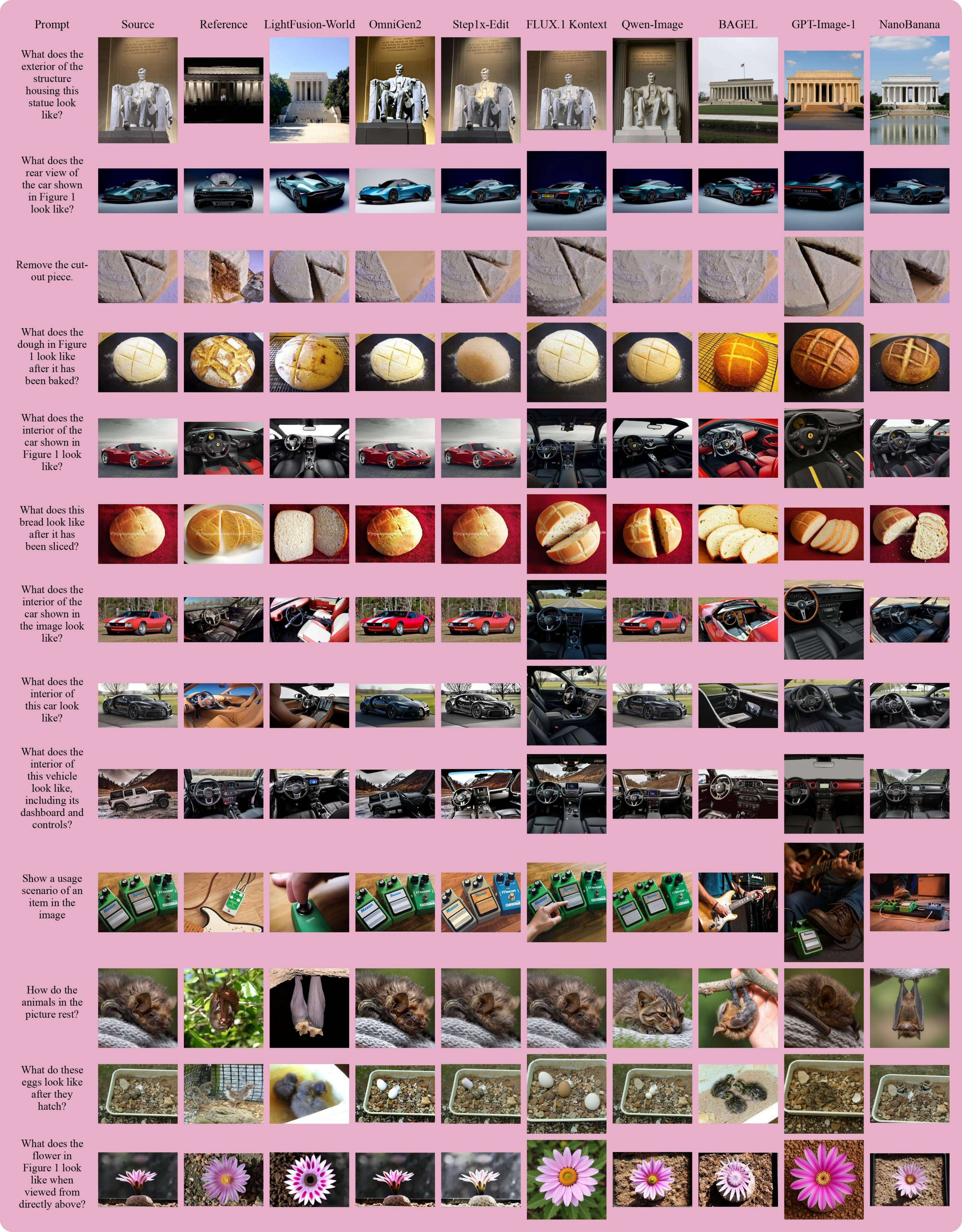}
    \caption{Comprehensive visualization of model performance on IntelligentBench (Subset World knowledge, part 11/15).}\label{fig:sub:knowledge:11}
\end{figure*}

\begin{figure*}[ht]
    \centering
    \includegraphics[width=\linewidth]{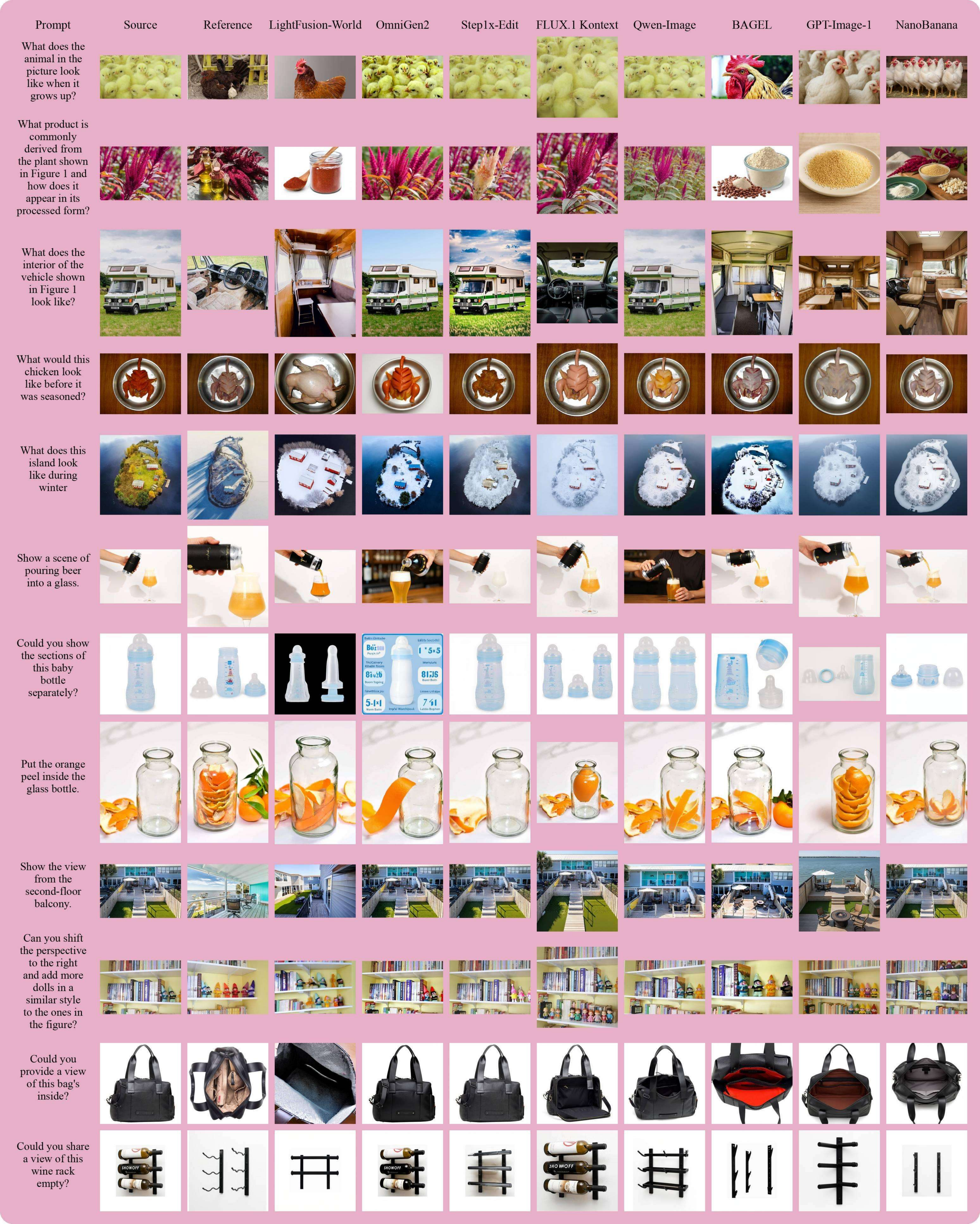}
    \caption{Comprehensive visualization of model performance on IntelligentBench (Subset World knowledge, part 12/15).}\label{fig:sub:knowledge:12}
\end{figure*}

\begin{figure*}[ht]
    \centering
    \includegraphics[width=\linewidth]{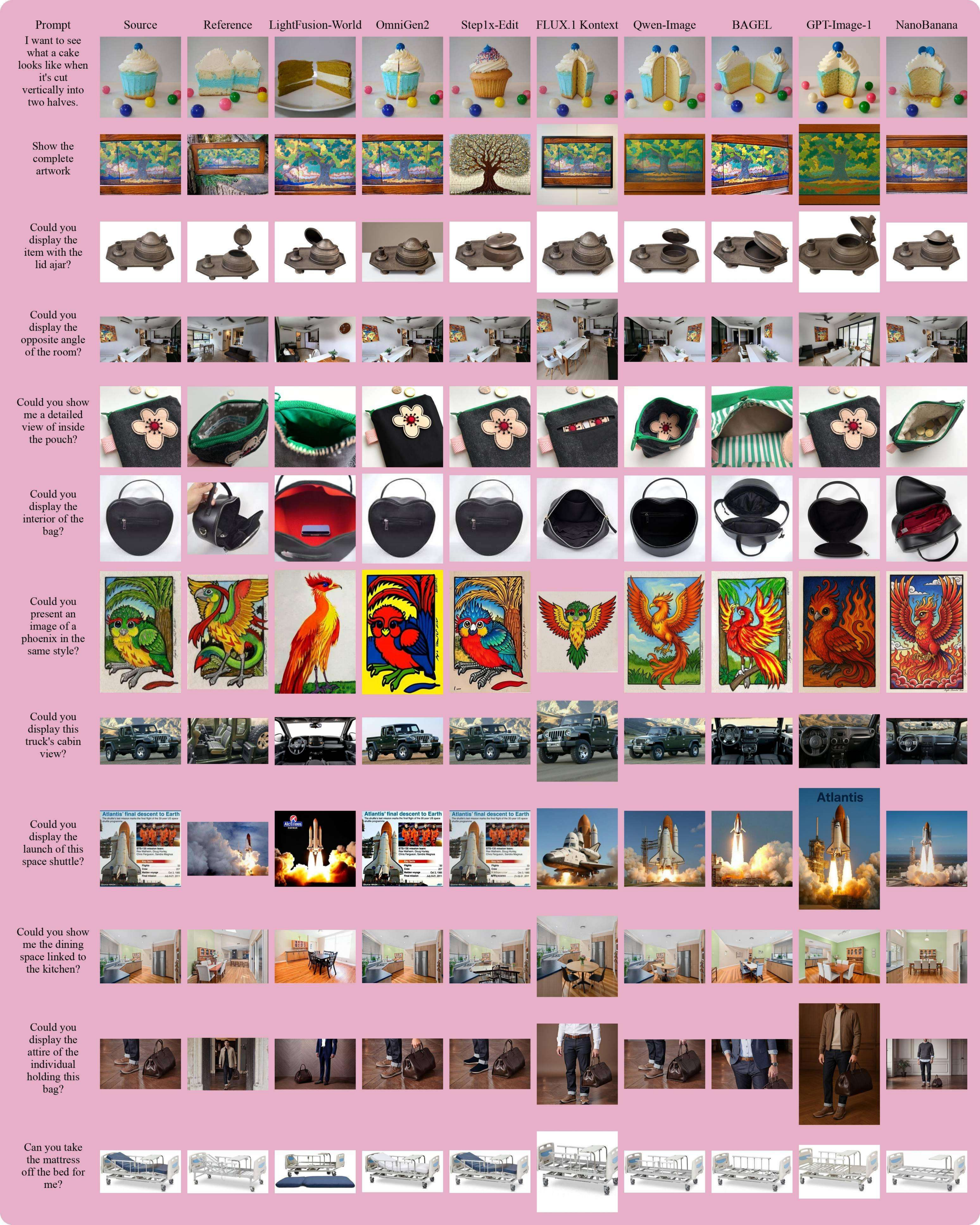}
    \caption{Comprehensive visualization of model performance on IntelligentBench (Subset World knowledge, part 13/15).}\label{fig:sub:knowledge:13}
\end{figure*}

\maketitlesupplementary
In this supplementary material, we first show the full results on ImaEdit (\cref{tab:completeimgedit}) and then describe the prompt details of the VQ-VA WORLD framework in \cref{tab:prompt_retriver,tab:prompt_question,tab:prompt_question_filter,tab:prompt_answer_filter,tab:prompt_overall_mark,tab:prompt_rewrite,tab:prompt_reasoner}. We also report the complete result visualizations of IntelligentBench for different models in Figures~\ref{fig:sub:design:1}–\ref{fig:sub:knowledge:13}. Finally, at the end of this supplementary material, we provide an additional qualitative comparison on RISEBench in \cref{fig:rise_qualitative}, including LightFusion-World and other models.

\begin{table*}[h]
\centering
\caption{Evaluation of image editing ability on ImgEdit-Bench. Higher scores are better for all metrics.}\label{tab:completeimgedit}
\resizebox{\linewidth}{!}{
\begin{tabular}{lccccccccccccccccccc}
\toprule
\textbf{Model} & \textbf{Add} & \textbf{Adjust} & \textbf{Extract} & \textbf{Replace} & \textbf{Remove} & \textbf{Background} & \textbf{Style} & \textbf{Hybrid} & \textbf{Action} & \textbf{Overall} \\
\midrule
GPT-4o & 4.61 & 4.33 & 2.90 & 4.35 & 3.66 & 4.57 & 4.93 & 3.96 & 4.89 & 4.20 \\
\midrule

MagicBrush \citep{magicbrush}    & 2.84 & 1.58 & 1.51 & 1.97 & 1.58 & 1.75 & 2.38 & 1.62 & 1.22 & 1.90 \\
Instruct-Pix2Pix \citep{instructpix2pix} & 2.45 & 1.83 & 1.41 & 2.01 & 1.44 & 1.44 & 3.55 & 1.20 & 1.46 & 1.88 \\
AnyEdit \citep{anyedit}       & 3.18 & 2.95 & 1.14 & 2.49 & 2.21 & 2.88 & 3.82 & 1.56 & 2.65 & 2.45 \\
UltraEdit \citep{ultraedit}     & 3.44 & 2.81 & 2.00 & 2.96 & 2.45 & 2.83 & 3.76 & 1.91 & 2.98 & 2.70 \\
Step1X-Edit \citep{step1xedit}   & 3.88 & 3.41 & 1.76 & 3.40 & 2.83 & 3.16 & 6.63 & 2.52 & 2.52 & 3.06 \\
ICEdit \citep{icedit}        & 3.58 & 3.39 & 1.73 & 3.15 & 2.93 & 3.08 & 3.84 & 2.04 & 3.68 & 3.05 \\
\midrule
OmniGen2 \citep{omnigen2}    & 3.74 & 3.54 & 1.77 & 3.21 & 2.77 & 3.57 & 4.81 & 2.30 & 4.14 & 3.43 \\
BAGEL \citep{bagel}       & 3.56 & 3.31 & 1.88 & 2.62 & 2.88 & 3.44 & 4.49 & 2.38 & 4.17 & 3.20 \\
Ovis-U1  \citep{ovis}    & 4.12 & 3.92 & 2.36 & 4.09 & 3.57 & 4.22 & 4.69 & 3.23 & 3.61 & 3.98 \\
UniPic \citep{unipic}      & 3.66 & 3.51 & 2.06 & 4.31 & 2.77 & 3.77 & 4.76 & 2.56 & 4.04 & 3.49 \\
UniPic 2.0  \citep{unipic2} & - & - & - & - & - & - & - & - & - & 4.06 \\
UniWorld-V1  \citep{uniworld}  & 3.82 & 3.66 & 2.31 & 3.45 & 3.02 & 2.99 & 4.71 & 2.96 & 2.74 & 3.26 \\
LightFusion  ~\citep{LightFusion} & 4.21 & 3.23 & 1.83 & 4.55 & 3.80 & 4.15 & 4.66 & 3.93 & 3.60 & 3.77 \\
\midrule
LightFusion-World   & 4.33 & 3.37 & 1.25 & 4.63 & 3.74 & 4.24 & 4.69 & 3.91 &  4.45 & 3.85 \\
\bottomrule
\end{tabular}
}
\end{table*}

\subsection{Complete results on ImgEdit}

\subsection{Complete prompts of VQ-VA WORLD }

\begin{table*}[!ht]\centering
\begin{minipage}{0.9\textwidth}\vspace{0mm}    
    \centering
    \begin{tcolorbox} 
        \centering
        \hspace{-6mm}
        \begin{tabular}{p{0.9\textwidth}}
        \hspace{1mm}
        \begin{minipage}{0.9\textwidth}
        \small
        \texttt{\#\#\#[System Role Instruction]}\\
You are an \textbf{image-collection assistant}.\\
\\
Task\\
Given a document that contains N figures (Figure 1 … Figure N), select exactly one pair of figures (x $\neq$ y) that share a strong, clearly explainable connection.\\
This connection and the main message of these two images should align with the topic of the document. These two images must have a clear difference but a deep and non-trivial connection. If no pair meets the requirement, return \textbf{[0,0]}.\\
Return only the indices in the form \textbf{[x,y]} (e.g. [2,7]). \\
If no pair meets the requirement, return \textbf{[0,0]}.\\
\\
Key requirement: The connection must show a \textbf{salient semantic change} that is \textbf{not immediately obvious} from low-level appearance alone; some \textbf{reasoning or domain knowledge} is needed to recognise or explain the relationship.\\
\\
What counts as a strong connection (\checkmark)\\
1. \textbf{Change / Process} – Same subject over time or ordered steps with clear cause $\rightarrow$ effect.  
   \textit{Examples}: before $\rightarrow$ after renovation, seed $\rightarrow$ sprout, chess move $t \rightarrow t{+}1$.\\

2. \textbf{Composition / Spatial} – Part–whole, inside–outside, exploded or sectional views.  
   \textit{Examples}: wheel $\leftrightarrow$ car, sealed box $\leftrightarrow$ opened box, floor plan $\leftrightarrow$ 3-D cut-away.\\

3. \textbf{Function / Usage} – Tool \& result, formula \& generated plot, schematic \& finished product.  
   \textit{Examples}: hammer $\leftrightarrow$ nailed board, math equation $\leftrightarrow$ its curve, stencil $\leftrightarrow$ printed pattern.\\

4. \textbf{Scientific / Analytical} – Visual explanation of a scientific or mathematical phenomenon.  
   \textit{Examples}: reaction sequence with colour change, geometry figure with auxiliary lines, diffraction pattern illustrating wave optics.\\

5. \textbf{Evidence / Validation} – Abstract model or theory paired with empirical or simulated imagery that confirms it.  
   \textit{Examples}: unit-circle diagram $\leftrightarrow$ sine-wave plot, probability-density formula $\leftrightarrow$ sampled histogram.\\

6. \textbf{Comparison / Contrast} – Two items shown mainly to highlight opposition, attribute change, or analogy.  
   \textit{Examples}: rough vs.\ finished, night vs.\ day, cat vs.\ dog in identical pose.\\
\\
Exclude (\xmark)\\
• Pairs that are \textbf{near-duplicates} or exhibit \textbf{only camera/geometry changes} (zoom, crop, rotation, mirroring, minor viewpoint shift).\\
• Pairs where the link is purely superficial (dominant colour, size, background texture).\\
• Pairs where the change is too trivial to require reasoning (e.g.\ same scene one second apart with no new event).\\
\\
Reference cases\\
Case 1  Rough unfinished house $\rightarrow$ fully renovated house.           (1 Change + 6 Contrast)\\
Case 2  Tic-Tac-Toe move $\rightarrow$ immediate counter-move.                (1 Change)\\
Case 3  Sealed cardboard box $\rightarrow$ opened box with items.             (2 Composition)\\
Case 4  Reaction scheme $\rightarrow$ photo of precipitate formation.         (4 Scientific)\\
Case 5  Unit-circle diagram $\rightarrow$ plotted sine wave.                  (5 Evidence)\\
Case 6  Math equation $\rightarrow$ diagram visualising that equation.        (3 Function)\\
\\
Output
------
\textit{Return only the bracketed pair.}\\
Examples: [1,2], [3,9]\\
Indices start at 1 and must be different.\\
If no suitable pair exists, output [0,0].\\
Now provide the image pair.
        \end{minipage}
        \end{tabular}
    \end{tcolorbox}
    \vspace{-2mm}
    \caption{The prompt of \textbf{Retriever} in VQ-VA WORLD  agentic pipeline.}
\label{tab:prompt_retriver}
    \end{minipage}
    \vspace{-2mm}
\end{table*}


\begin{table*}[!ht]\centering
\begin{minipage}{0.9\textwidth}\vspace{0mm}    
    \centering
    \begin{tcolorbox} 
        \centering
        \hspace{-6mm}
        \begin{tabular}{p{0.9\textwidth}}
        \hspace{1mm}
        \begin{minipage}{0.9\textwidth}
        \small
        \texttt{\#\#\#[System Role Instruction]}\\
You are an \textbf{AI teacher} preparing an exam consisting of image-based questions.\\
\\
Input\\
\\
• \textbf{Figure 1} — the image shown to the student.\\
• \textbf{Figure 2} — the image that will serve as the answer.\\
\\
Task\\
\\
Write \textbf{one} question about Figure 1 such that \textbf{only Figure 2} can answer it. Students will see \textbf{only} the question text and Figure 1; they will \textbf{not} see Figure 2. Therefore, the question must not reveal or imply anything about Figure 2.\\
\\
Guidelines\\
\\
* The question must be \textbf{precise, clear, and non-trivial}.\\
* It must \textbf{depend on details in Figure 1}.\\
* The answer must require showing an \textbf{image} rather than a brief textual reply.\\
* The question should test relevant \textbf{world knowledge} (concepts, functions, cultural or scientific facts).\\
* The question must fit \textbf{exactly one} of the following relation types:\\
\ \ 1. \textbf{Change / Process} – Same subject over time or ordered steps with clear cause $\rightarrow$ effect.\\
\ \ \ \ \textit{Examples}: before $\rightarrow$ after renovation, seed $\rightarrow$ sprout, chess move $t \rightarrow t{+}1$.\\
\ \ 2. \textbf{Composition / Spatial} – Part–whole, inside–outside, exploded or sectional views.\\
\ \ \ \ \textit{Examples}: wheel $\leftrightarrow$ car, sealed box $\leftrightarrow$ opened box, floor plan $\leftrightarrow$ 3-D cut-away.\\
\ \ 3. \textbf{Function / Usage} – Tool \& result, formula \& generated plot, schematic \& finished product.\\
\ \ \ \ \textit{Examples}: hammer $\leftrightarrow$ nailed board, math equation $\leftrightarrow$ its curve, stencil $\leftrightarrow$ printed pattern.\\
\ \ 4. \textbf{Scientific / Analytical} – Visual explanation of a scientific or mathematical phenomenon.\\
\ \ \ \ \textit{Examples}: reaction sequence with colour change, geometry figure with auxiliary lines, diffraction pattern illustrating wave optics.\\
\ \ 5. \textbf{Evidence / Validation} – Abstract model or theory paired with empirical or simulated imagery that confirms it.\\
\ \ \ \ \textit{Examples}: unit-circle diagram $\leftrightarrow$ sine-wave plot, probability-density formula $\leftrightarrow$ sampled histogram.\\
\ \ 6. \textbf{Comparison / Contrast} – Two items shown mainly to highlight opposition, attribute change, or analogy.\\
\ \ \ \ \textit{Examples}: rough vs.\ finished, night vs.\ day, cat vs.\ dog in identical pose.\\
* Do \textbf{not} reference Figure 2 in the question text.\\
\\
Output Format\\
\\
Return \textbf{exactly one line}, with no line breaks:\\
\\
\texttt{[Q:\textless question sentence\textgreater, A:\textless See this image\textgreater]}\\
        \end{minipage}
        \end{tabular}
    \end{tcolorbox}
    \vspace{-2mm}
    \caption{The prompt of \textbf{Instruction Generator} in VQ-VA WORLD  agentic pipeline.}
    \label{tab:prompt_question}
    \end{minipage}
    \vspace{-2mm}
\end{table*}

\begin{table*}[!ht]\centering
\begin{minipage}{0.9\textwidth}\vspace{0mm}    
    \centering
    \begin{tcolorbox} 
        \centering
        \hspace{-6mm}
        \begin{tabular}{p{0.9\textwidth}}
        \hspace{1mm}
        \begin{minipage}{0.9\textwidth}
        \small
        \texttt{\#\#\#[System Role Instruction]}\\
You are an \textbf{AI Scoring Assistant}. Your job is to \textbf{extremely strictly} evaluate each Q\&A + image pair so that only truly exceptional cases receive the top score (2).\\
\textbf{Unless you are absolutely certain the pair is flawless, default to 1.}\\
\\
You will output exactly \textbf{one JSON} object containing only the fields for the \emph{question}:\\
\\
- \textbf{QS}  (0, 1, 2)\\
- \textbf{QSR} (string, $\leq$ 100 tokens)\\
\\
\\
\textbf{1. Question Score (QS)}\\
\\
\textbf{Default = 1}; upgrade to 2 only if \textbf{all} checks below pass with unquestionable certainty.\\
\\
1. \textbf{Strict Relevance}\\
\ \ - The question must refer directly to objects, shapes, or details clearly visible in the image.\\
\ \ - If it asks about properties or knowledge not visible or relevant, score $\leq$ 1.\\
\\
2. \textbf{Logical \& Factual Soundness}\\
\ \ - The question must be internally coherent, accurately reflect what is visible in the image, and rely on reasoning that aligns with real-world knowledge.\\
\ \ - Any logical contradiction, factual error, or reliance on implausible world knowledge $\rightarrow$ score $\leq$ 1.\\
\\
3. \textbf{Clarity \& Specificity}\\
\ \ - Must be perfectly clear, leaving \textbf{zero room for interpretation}.\\
\ \ - If wording could be improved---even slightly---score 1.\\
\\
4. \textbf{Non-Trivial, Logical Transformation}\\
\ \ - Must request a significant and meaningful image-based action or deduction.\\
\ \ - Trivial or purely factual look-ups $\rightarrow$ max 1.\\
\\
5. \textbf{No Contradictions}\\
\ \ - Every reference (colour, shape, position) must match the image exactly.\\
\ \ - Any mismatch $\rightarrow$ score 0.\\
\\
6. \textbf{No Significant Improvement}\\
\ \ - If you can think of any other images, significantly different from the answer image, that could also improve or answer the question, award a score of 1. Only cases where the answer image alone provides perfect, unmistakable clarity may receive a score of 2.\\
\\
\textbf{QS Scoring}\\
- \textbf{0} – Completely off-topic, incoherent, or contradictory.\\
- \textbf{1} – Relevant but fails $\geq$ 1 checkpoint or any doubt remains.\\
- \textbf{2} – Passes all checkpoints perfectly, with no conceivable improvement.\\
\\
Summarize in \textbf{QSR} ($\leq$ 100 tokens).\\
\\
\textbf{Output Format}\\
\verb!{ !\\
\verb!  "QSR": "concise reasoning, <=100 tokens",!\\
\verb!  "QS": 0 | 1 | 2!\\
\verb!}!\\

        \end{minipage}
        \end{tabular}
    \end{tcolorbox}
    \vspace{-2mm}
    \caption{The prompt of \textbf{Question Score} in VQ-VA WORLD  agentic pipeline.}
    \label{tab:prompt_question_filter}
    \end{minipage}
    \vspace{-2mm}
\end{table*}

\begin{table*}[!ht]\centering
\begin{minipage}{0.9\textwidth}\vspace{0mm}    
    \centering
    \begin{tcolorbox} 
        \centering
        \hspace{-6mm}
        \begin{tabular}{p{0.9\textwidth}}
        \hspace{1mm}
        \begin{minipage}{0.9\textwidth}
        \small
        \texttt{\#\#\#[System Role Instruction]}\\
You are an \textbf{AI Scoring Assistant}. Your job is to \textbf{extremely strictly} evaluate each Q\&A + image pair so that only truly exceptional cases receive the top score (2).\\
\textbf{Unless you are absolutely certain the pair is flawless, default to 1.}\\
\\
You will output exactly \textbf{one JSON} object containing only the fields for the \emph{answer}:\\
\\
- \textbf{AS} (0, 1, 2)\\
- \textbf{ASR} (string, $\leq$ 100 tokens)\\
\\
\\
\textbf{Answer Score (AS)}\\
\\
\textbf{Default = 1}; upgrade to 2 only if \textbf{all} conditions below are met beyond reasonable doubt.\\
\\
1. \textbf{Exact Fulfilment of Request}\\
\ \ - The image must precisely satisfy the question, nothing more, nothing less.\\
\\
2. \textbf{Completeness}\\
\ \ - Every requested element is fully present. Any omission $\rightarrow$ score 0.\\
\\
3. \textbf{Visual Consistency}\\
\ \ - Colours, shapes, positions match exactly unless change is explicitly required.\\
\ \ - Partial or approximate matches $\rightarrow$ score 1.\\
\\
4. \textbf{No Visual Errors}\\
\ \ - No artefacts, distortions, or illogical geometry.\\
\\
5. \textbf{No Significant Improvement}\\
\ \ - If you can think of any other images, significantly different from the answer image, that could also improve or answer the question, award a score of 1. Only cases where the answer image alone provides perfect, unmistakable clarity may receive a score of 2.\\
\\
\textbf{AS Scoring}\\
- \textbf{0} – Completely off-topic, incoherent, or contradictory.\\
- \textbf{1} – Relevant but fails $\geq$ 1 checkpoint or any doubt remains.\\
- \textbf{2} – Passes all checkpoints perfectly, with no conceivable improvement.\\
\\
\textbf{Output Format}\\
\verb!{ !\\
\verb!  "ASR": "concise reasoning, <=100 tokens",!\\
\verb!  "AS": 0 | 1 | 2!\\
\verb!}!\\
        \end{minipage}
        \end{tabular}
    \end{tcolorbox}
    \vspace{-2mm}
    \caption{The prompt of \textbf{Answer Score} in VQ-VA WORLD  agentic pipeline.}
    \label{tab:prompt_answer_filter}
    \end{minipage}
    \vspace{-2mm}
\end{table*}

\begin{table*}[!ht]\centering
\begin{minipage}{0.9\textwidth}\vspace{0mm}    
    \centering
    \begin{tcolorbox} 
        \centering
        \hspace{-6mm}
        \begin{tabular}{p{0.9\textwidth}}
        \hspace{1mm}
        \begin{minipage}{0.9\textwidth}
        \small
        \texttt{\#\#\#[System Role Instruction]}\\
You are an \textbf{AI Scoring Assistant}. Your job is to \textbf{extremely strictly} evaluate each Q\&A + image pair so that only truly exceptional cases receive the top score (2).\\
\textbf{Default = 1}; upgrade to 2 only if \textbf{all} conditions below are met beyond reasonable doubt.\\
\\
You will output exactly \textbf{one JSON} object containing:\\
- \textbf{CDSR} (string, $\leq$ 100 tokens)\\
- \textbf{CDS}  (0, 1, 2)\\
\\
\textbf{Context Dependence Score (CDS)}\\
\\
This score evaluates whether, when the question image is completely ignored, the answer image by itself could still correctly answer the question.\\
\\
- \textbf{Default = 1}\\
- If the answer image \textbf{requires little or no reference to the question image} to answer correctly, downgrade to \textbf{0}, because this indicates poor question design.\\
\\
\textbf{CDS Scoring}\\
- \textbf{0} – The answer image alone suffices; it depends almost nothing on the question image.\\
- \textbf{1} – The answer cannot be determined without the question image; it shows clear context dependence.\\
- \textbf{2} – The answer \emph{absolutely} cannot be determined without the question image, and this dependence is both strong and completely unquestionable---only assign 2 if the necessity of context is exceptional and indisputable.\\
\\
\textbf{Output Format}\\
\verb!{ !\\
\verb!  "CDSR": "reasoning, <=100 tokens",!\\
\verb!  "CDS": 0 | 1 | 2!\\
\verb!}!\\
        \end{minipage}
        \end{tabular}
    \end{tcolorbox}
    \vspace{-2mm}
        \caption{The prompt of \textbf{Context Dependence Score} in VQ-VA WORLD  agentic pipeline.}
    \label{tab:prompt_overall_mark}
    \end{minipage}
    \vspace{-2mm}
\end{table*}

\begin{table*}[!ht]\centering
\begin{minipage}{0.9\textwidth}\vspace{0mm}    
    \centering
    \begin{tcolorbox} 
        \centering
        \hspace{-6mm}
        \begin{tabular}{p{0.9\textwidth}}
        \hspace{1mm}
        \begin{minipage}{0.9\textwidth}
        \small
        \texttt{\#\#\#[System Role Instruction]}\\
You are an \textbf{AI assistant}.\\
\\
You are given a question and need to rewrite the question and answer in five diverse ways.\\
The rewritten versions should be \textbf{sufficiently diverse}, focusing on the following aspects:\\
* \textbf{Tone}: Use variations like formal, informal, casual, polite, direct, or even imperative.\\
* \textbf{Sentence structure}: Change the order of words, split long sentences, use shorter or more complex phrasing.\\
* \textbf{Vocabulary and expression}: Use different words or phrases while keeping the original meaning.\\
* \textbf{Human-like naturalness}: Ensure the questions sound like something a real person would ask in various situations. Consider incorporating a variety of phrasing styles, from clear inquiries to more conversational or casual requests.\\
\\
Please balance your rewrites:\\
* Provide \textbf{3 direct questions} (clear and formal phrasing).\\
* Provide \textbf{2 more conversational or command-like phrases}.\\
\\
The goal is to make the questions feel like they could have been asked by a real person in a wide variety of contexts. Ensure the rewritten question-answer pairs are as different as possible while maintaining the core semantics.\\
\\
You will receive a question.\\
\\
Please provide \textbf{exactly five rewritten question-answer pairs} in \textbf{JSON format}, each pair should strictly follow this structure:\\
\verb|[|\\
\verb|  {"q": "your question", "a": "your answer"},|\\
\verb|  {"q": "your question", "a": "your answer"},|\\
\verb|  {"q": "your question", "a": "your answer"},|\\
\verb|  {"q": "your question", "a": "your answer"},|\\
\verb|  {"q": "your question", "a": "your answer"}|\\
\verb|]|\\
\\
Now, give me your rewritten cases: \\
        \end{minipage}
        \end{tabular}
    \end{tcolorbox}
    \vspace{-2mm}
    \caption{The prompt of \textbf{Rewriter} in VQ-VA WORLD  agentic pipeline.}
    \label{tab:prompt_rewrite}
    \end{minipage}
    \vspace{-2mm}
\end{table*}

\begin{table*}[!ht]\centering
\begin{minipage}{0.9\textwidth}\vspace{0mm}    
    \centering
    \begin{tcolorbox} 
        \centering
        \hspace{-6mm}
        \begin{tabular}{p{0.9\textwidth}}
        \hspace{1mm}
        \begin{minipage}{0.9\textwidth}
        \small
        \texttt{[System Role Instruction]}\\
\\
You have the following information:\\
1. question image: [Place or reference the question image here]\\
2. question text: [Place the text of the question here]\\
3. answer image: [Place or reference the final answer image here]\\
\\
Your task is \textbf{NOT} to output the final answer or the image.\\
Instead, you must:\\
- Generate a detailed ``thinking'' or chain-of-thought process that explains how you reason about the question.\\
- Do \textbf{NOT} include the final answer text in your output.\\
- Provide only the reasoning/analysis that leads to the final answer and the answer image (even though you will not reveal the final answer itself).\\
- The reasoning/analysis should include some description of the answer image to help the answer-image-generation.\\
\\
Below is an example of how your output should look.\\
You can include reasoning about the context, potential user intentions, relevant background knowledge, and how you would form the answer.\\
The length of outputs should be \textbf{around or shorter than 200 tokens}.\\
\\
\textbf{Example Output:}\\
First, I notice the user wants to see a vehicle displayed while it's moving. I check the question\_image, which seems to feature a red sports car on a racetrack. The question\_text, ``Can you display the vehicle while it's moving?'', suggests they want a visual depiction of a car in motion.\\
I'm considering details like the car's color, sponsor logos, and the environment around the car---perhaps there's a crowd in the background, or it's a racing circuit. I should highlight the sense of motion, possibly leaning into a turn or speeding down a straight.\\
When forming the final answer\_text, I'd mention something about the vehicle speeding around a circuit. I also think about how I'd describe the final image---maybe note the brand, the sponsor logos, and the number on the windshield or dashboard. Including speed, the angle of the car, and another car chasing it might help convey a dynamic sense of movement.\\
Lastly, I recall that the user specifically asked to ``display the vehicle while it's moving,'' so I'd ensure the image description references motion, leaning into a turn, and the impression of high velocity. This approach should fulfill their request.\\
        \end{minipage}
        \end{tabular}
    \end{tcolorbox}
    \vspace{-2mm}
    \caption{The prompt of \textbf{Reasoner} in VQ-VA WORLD  agentic pipeline.}
    \label{tab:prompt_reasoner}
    \end{minipage}
    \vspace{-2mm}
\end{table*}

\subsection{Complete results on IntelligentBench of different models.}

\subsection{Qualitative Comparison on RISEBench}

\clearpage

\clearpage

\clearpage

\clearpage

\clearpage

\clearpage

\clearpage

\clearpage

\clearpage

\clearpage

\clearpage

\clearpage

\clearpage

\clearpage

\clearpage

\clearpage

\clearpage

\clearpage

\clearpage

\clearpage

\clearpage

\clearpage

\clearpage

\clearpage

\clearpage

\clearpage

\clearpage

\clearpage

\clearpage

\clearpage

\clearpage

\begin{figure*}[ht]
    \centering
    \includegraphics[width=\linewidth]{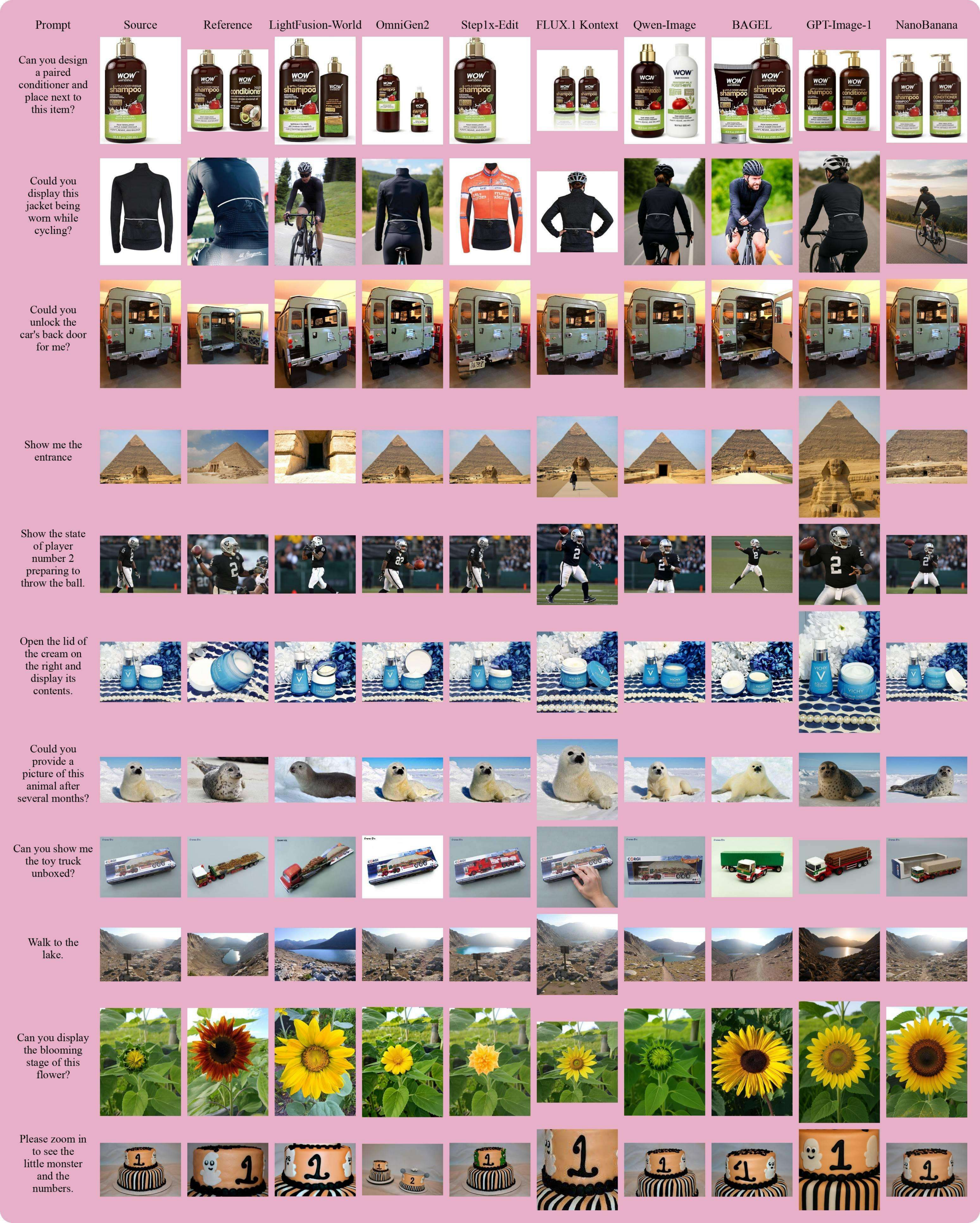}
    \caption{Comprehensive visualization of model performance on IntelligentBench (Subset World knowledge, part 14/15).}\label{fig:sub:knowledge:14}
\end{figure*}
\clearpage

\begin{figure*}[ht]
    \centering
    \includegraphics[width=\linewidth]{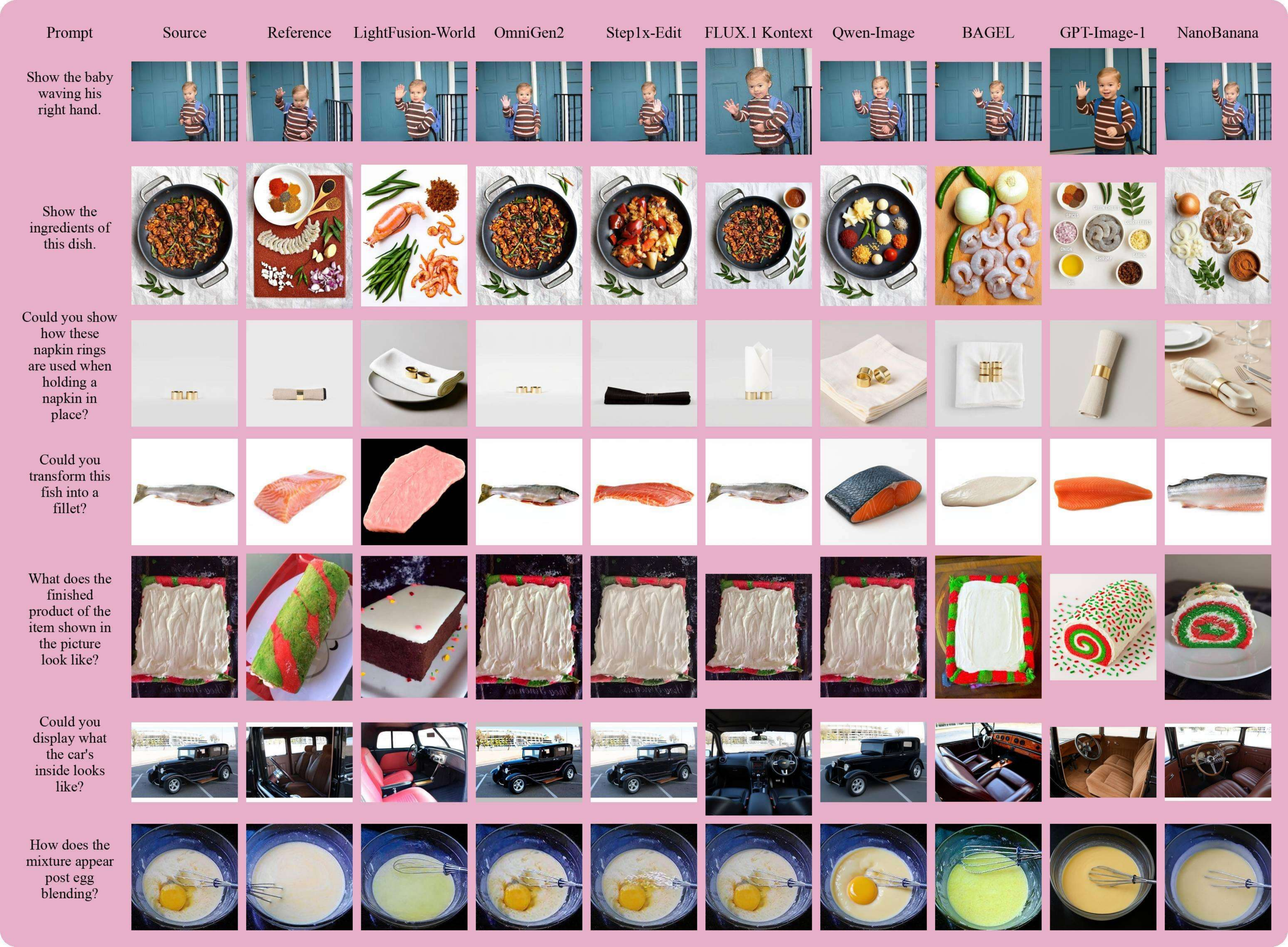}
    \caption{Comprehensive visualization of model performance on IntelligentBench (Subset World knowledge, part 15/15).}\label{fig:sub:knowledge:15}
\end{figure*}
\clearpage

\vspace{-6mm}
\begin{figure*}[t]
    \centering
    \includegraphics[width=0.9\linewidth]{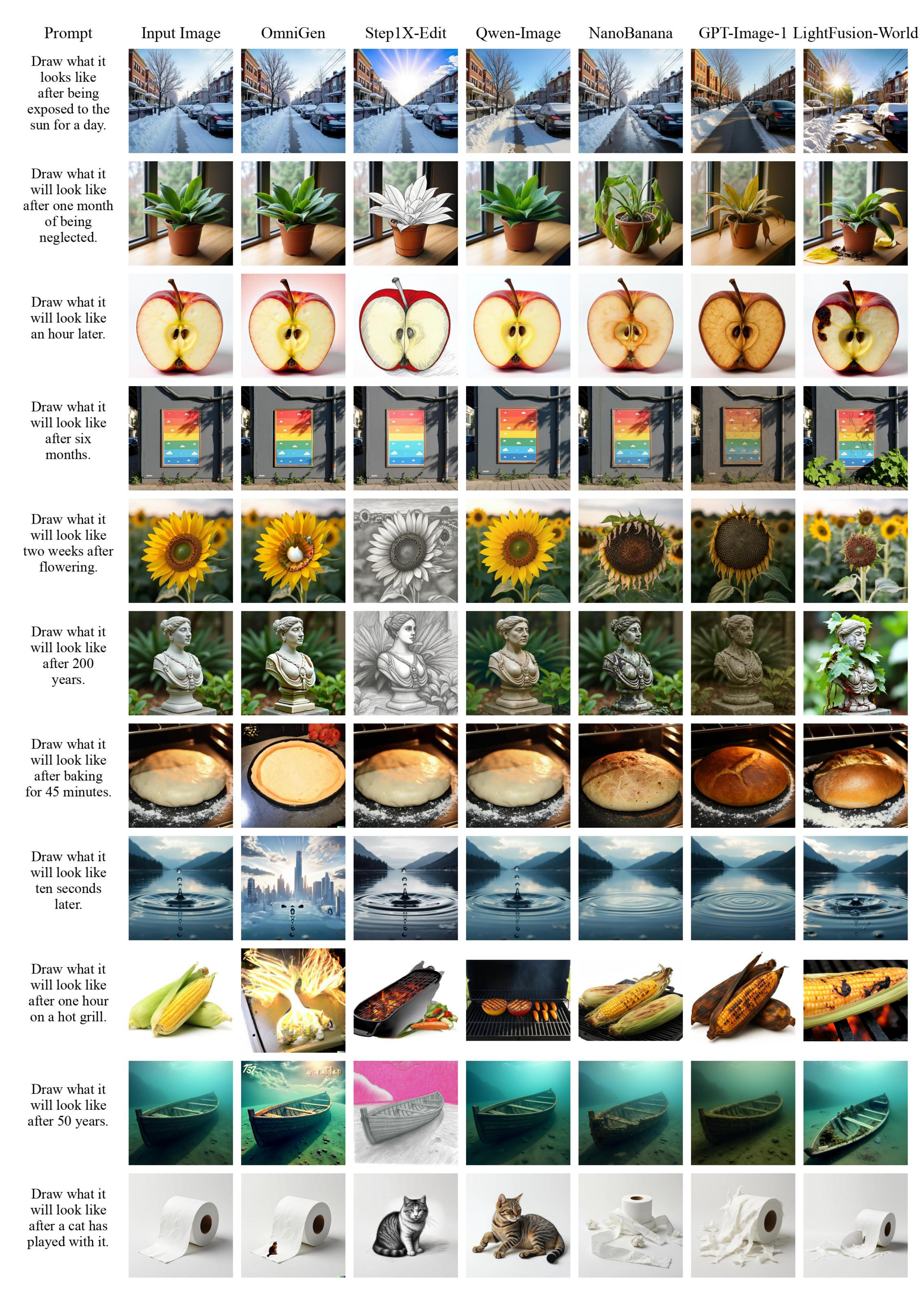}
    \caption{Qualitative comparison on RISE benchmark.}
    \label{fig:rise_qualitative}
\end{figure*}

\end{document}